\newcommand{\ac}[1]{\textcolor{magenta}{\sf {\bf AC:}  #1}} 

\newcommand{\hs}[1]{\textcolor{orange}{\sf {\bf HS:}  #1}}
\newcommand{\gd}[1]{\textcolor{blue}{\sf {\bf GD:}  #1}}

\documentclass{ws-book9x6}

\usepackage{ws-book-thm}   
\usepackage{ws-book-har}   
\usepackage[pdfpagelabels=false]{hyperref}  
\usepackage{algpseudocode}
 \usepackage{booktabs}
 \usepackage{subcaption}
 \usepackage{algorithm}
\usepackage{amsmath}
\usepackage{amssymb}
\usepackage{multirow}
\usepackage{todonotes}
\usepackage[title]{appendix}
\title{}      

\begin{document}


\setcounter{page}{1}

\chapter{Drones that Think on their Feet: Sudden Landing Decisions with Embodied AI}

\author{Diego Ortiz$^\dag$ Mohit Agrawal$^\dag$ Yash Malegaonkar$^\dag$ Luis Burbano$^\dag$ \\Axel Andersson$^\ddag$ György Dán$^\ddag$ Henrik Sandberg$^\ddag$ Alvaro A. Cardenas$^\dag$
\\$^\dag$ University of California, Santa Cruz\\
$^\ddag$KTH  Royal Institute of Technology}

    Autonomous drones must often respond to sudden events, such as alarms, faults, or unexpected changes in their environment, that require immediate and adaptive decision-making. Traditional approaches rely on safety engineers' hand-coding large sets of recovery rules, but this strategy cannot anticipate the vast range of real-world contingencies and quickly becomes incomplete. Recent advances in embodied AI, powered by large visual-language models, provide common-sense reasoning to assess context and generate appropriate actions in real time. We demonstrate this capability in a simulated urban benchmark in the Unreal Engine, where drones dynamically interpret their surroundings and decide on sudden maneuvers for safe landings. Our results show that embodied AI makes possible a new class of adaptive recovery and decision-making pipelines that were previously infeasible to design by hand, advancing resilience and safety in autonomous aerial systems.

\section{Introduction}\label{sec1.1}

Autonomous drones hold enormous promise for society, with potential applications in disaster response, infrastructure inspection, environmental monitoring, and transportation. Their ability to operate in dynamic environments and carry out missions without constant human supervision could transform how critical services are delivered. However, to fulfill this promise, drones must be trustworthy. In particular, they must be prepared to respond when alarms or unexpected events occur. These alarms may signal cyber attacks such as GPS spoofing, electromagnetic interference (EMI) signal injection~\cite{jang2023ndss}, device failures such as sensor degradation, or adverse conditions such as sudden weather changes or unexpected obstacles. Although rare, such events can have severe consequences, and autonomous agents must be able to assess the situation and determine an appropriate course of action in real time.

Previous work recognizes the need for automatic recovery when alarms or unexpected events occur, but existing approaches are tailored to \emph{specific} failure models and lack general, context-aware reasoning~\cite{cardenas2025response,barbosa2025robust}. \textbf{(A) Recovery to pre-specified targets.} A popular recovery method formalizes how to steer a system to a \emph{pre-defined} safe set after an alert~\cite{zhang2020rtss,zhang2024fast}. These approaches assume that the target (safe) set is known in advance and valid throughout recovery; they do not revise that target if the environment becomes unsafe at runtime. \textbf{(B) Mitigation based on experience.} Complementary lines address particular attack or failure classes by learning from previous successful traces~\cite{dash2021pidpiper}; however, if the failure condition requires a completely new maneuver or destination not seen before, these approaches may struggle. In general, previous efforts are \emph{scenario-bound}: they depend on pre-enumerated fault/attack models, static safe-target definitions, or fixed fallback behaviors, rather than dynamically \emph{interpreting} complex scenes and choosing actions with common-sense reasoning in open-ended, evolving environments.

In this work, we introduce a new pipeline that leverages large visual-language models (LVLMs) to support real-time decision making when sudden events occur. Our approach focuses on the case of sudden landing maneuvers, where a drone must quickly assess its surroundings and select a safe course of action. The pipeline integrates traditional control modules with LVLM-based reasoning: perception modules identify candidate surfaces, the LVLM evaluates their suitability using common-sense reasoning, and a movement planner executes the maneuver. To evaluate this approach, we build a benchmark on top of open source platforms, leveraging the Unreal Engine to propose realistic and dynamic-looking environments, with the Cosys-AirSim simulator, modeling the sensors and control stack of drones. 

This benchmark provides diverse, realistic urban scenarios with dynamic obstacles and configurable conditions, enabling systematic testing of recovery pipelines. Our contributions are threefold: 
\begin{enumerate}
    \item  We propose an LVLM-driven pipeline for sudden landing decisions that combines perception, reasoning, and control;
    \item We develop a reproducible benchmark for evaluating such pipelines in realistic urban simulations. Our benchmark is openly accessible at \url{https://github.com/RollingBeatle/Airsim-closeloop}
    \item We demonstrate through experiments that embodied AI enables adaptive recovery behaviors that were previously infeasible.
    \item We show the capabilities of out-of-the-box LVLMs to identify and reason over safety concerns and accomplish a safe landing with acceptable latency rates
    
\end{enumerate}




\section{Related Work}
Early work in mission recovery for autonomous systems often relies on the assumption that the region to recover to is known beforehand. Authors in~\cite{dash2021pidpiper} implicitly assume that the original target location is safe to go to regardless of the detected anomalies, and propose a controller based on the simplex architecture, where the recovery controller is activated under adverse conditions. The recovery controller is a neural network trained to mimic the behavior of a PID controller. A similar idea is presented in~\cite{dash2024specguard}, where a reinforcement learning policy is learned to satisfy safety constraints specified using signal temporal logic under adverse conditions. Focusing on recovery from sensor attacks, the implicit assumption is that the originally intended target location is safe. In~\cite{roque2022corridormpc}, a safe model-predictive control problem is formulated by using control barrier functions, which is similar to~\cite{dash2024specguard} and~\cite{zhang2024fast} in that safe regions and constraints are formulated beforehand. 

Safe landing zone (SLZ) detection is an active area of research. Current methods rely on geometric and semantic analysis of onboard perception from RGB-cameras and/or LiDAR. These methods leverage techniques from classical image analysis, machine learning, and deep learning. In \cite{maturana3dcnnforlzd}, the authors train a 3D convolutional neural network (CNN) to detect SLZs in rural areas for autonomous helicopters from a LiDAR point cloud. \cite{leelandingarearecognitiondl} uses a 2D CNN to find designated landing zones (helipads) from camera images with a feature matching algorithm. These deep learning based approaches rely on labeled data for supervised training of the models; hence, they would perform best in domains similar to those where the training data were collected. Complementary to these learning-based methods, \cite{yangmonocularvisionslam} proposes a GPS-denied monocular-vision SLAM pipeline that reconstructs a 3D point cloud and transforms it into a height-annotated grid to find flat landing zones. The paper \cite{safeuavmarcualina} formulates safe landing zone (SLZ) detection as a semantic segmentation task. Using a semi-synthetic dataset of aerial imagery, the authors train a U-Net model to assign each pixel to one of three categories: horizontal, vertical, or other. Candidate landing areas are then selected based on the assumption that pixels classified as horizontal correspond to locally planar terrain suitable for landing. \cite{delatorrevanegas2025visionbasedriskawareemergency} estimates SLZs by constructing an image-space risk map from the onboard camera stream. A segmentation module identifies object classes, each of which is assigned a risk score reflecting its presumed landing hazard. The resulting per-pixel risk map is then minimized to select a low-risk landing location in the image plane, which is used as the target landing position.

A survey on SLZ detection \cite{shahalamslzsurvey} highlights several open directions. One of them is to improve terrain-awareness. The surface of water is often flat and might seem like a viable landing option to an algorithm that successfully detects flat areas, but is unaware of the terrain. Another direction suggested in the survey is to couple SLZ localization with path planning and obstacle avoidance.

Foundation Models (FMs) have recently been integrated into various robotic systems for their reasoning capabilities. End-to-end models, such as Gemini Robotics~\cite{team2025geminigeminirobotics}, use a Vision-Language-Action (VLA) model that takes instructions in natural language and outputs low-level actuator commands in order to complete the task. Compound models combine FMs and classical control algorithms such as PID or model-predictive controllers. In \cite{sinha2024realtimeanomalydetectionreactive}, the authors use FMs in a drone to detect anomalies during flight and to select an appropriate region from a pre-defined set of regions. A VLM is used to convert camera images from the drone to text that determines which recovery region is selected. The main shortcoming of such approaches is that training and inference using such models is computationally intensive. End-to-end solutions for autonomous cars are starting to appear in the literature. In \cite{xiao2022endtoenddriving} the authors use conditional imitation learning to predict steering angle, throttle and brake actions from raw LiDAR and camera data. They investigate different methods to fuse these modalities in neural networks. \cite{prakash2021multimodaltransformer} leverages the transformer architecture to predict suitable way points from LiDAR and camera data, but offloads the task to low-level control to traditional PID controllers.  

A different line of work provides formal guarantees of recovery to pre-defined regions using control-theoretic approaches~\cite{zhang2024fast, zhangrealtimedatapredictive,leudo2023observer,garg2022control,zhangattackrecoverylqr}. These works primarily focus on recovering control systems, such as drones, from sensor faults or attacks. The issue is that it is not always clear how to define safe regions in practice, as in a dynamic environment, the safety of a region may change over time, e.g., due to moving obstacles.

In addition to computer-vision and control-oriented works, there exist decision support and situational awareness systems that reason over dynamic environments and safety constraints, even if they do not use visual reasoning for landing. For example, Insaurralde et al.~\cite{insaurralde2022ontology} builds an ontology-driven decision support system that evaluates hazards in airspace (e.g. weather, traffic, changing surveillance inputs) and helps determine whether operations (like takeoff or routing) remain safe in real time. 

Contrary to previous work, in this work, we explore how to leverage the reasoning capabilities of FMs for planning emergency landings in diverse realistic scenarios. We adopt a compound model, i.e., low-level control is via a traditional controller (such as a PID controller or MPC), which keeps compute requirements moderate. Conceptually, the FM assumes the role of a human operator who, in the event of an emergency, selects a safe landing spot based on visual depth information. Safety in this context means that the drone lands in a flat area without humans or obstacles nearby. 


\section{Assumptions and Problem Formulation}

\textbf{Assumptions.} Modern drones are equipped with a variety of sensors that provide complementary views of the environment. Typical platforms carry RGB cameras (monocular or stereo), depth sensors or LiDAR, GPS units, and sometimes radar altimeters or thermal cameras. For the scope of this work, we assume the drone is equipped with a downward-facing RGB camera, a basic LiDAR, and state-estimation sensors (e.g., IMU). These assumptions are realistic for both commercial and research platforms, while keeping the sensing payload lightweight enough for small UAVs operating in urban environments. We also assume that the camera, LiDAR, and actuators (in addition to our pipeline) are trustworthy during the recovery maneuver. 

\textbf{Problem formulation.} We consider the situation in which a drone receives an alarm or experiences an unexpected event (e.g., an attack, a device failure, or a sudden environmental change). The drone must respond by identifying a feasible recovery maneuver, such as a sudden landing, that ensures safety. Formally, let the environment at time $t$ be represented by sensor observations $O_t$ (images, depth estimates, or state information). Given an alert event $E_t$, the recovery problem is to compute a control sequence that drives the drone from its current state to a safe area in a short time. We informally define a safe state as one in which the drone has landed on a valid surface that avoids imminent hazards. The challenge is that safe areas are not known a priori and may evolve dynamically due to moving obstacles or environmental changes. Thus, the agent must infer both (i) which regions of the environment are plausible candidates for recovery and (ii) which sequence of actions will reliably reach them.

\textbf{Challenges.} The problem formulation raises several challenges. (1) \textbf{Uncertainty in perception:} vision- and depth-based modules may misclassify surfaces due to clutter, reflections, or sensor noise. (2) \textbf{Dynamic environments:} a region judged safe at time $t$ may become unsafe by $t+T$ due to pedestrians, traffic, or adverse weather. (3) \textbf{Reasoning under novelty:} alerts may correspond to scenarios never observed during training, requiring the system to extrapolate beyond predefined models. (4) \textbf{Real-time constraints:} all inference and planning must be completed quickly enough to be actionable on a resource-limited aerial platform.

While classical emergency recovery methods assume fixed safe sets or pre-enumerated failure modes, our objective is to enable drones to dynamically interpret their environment and compute recovery actions in real time, even when alarms reveal novel or unforeseen situations.

\begin{figure}[ht]
    \centering  
         \includegraphics[width=.7\columnwidth]{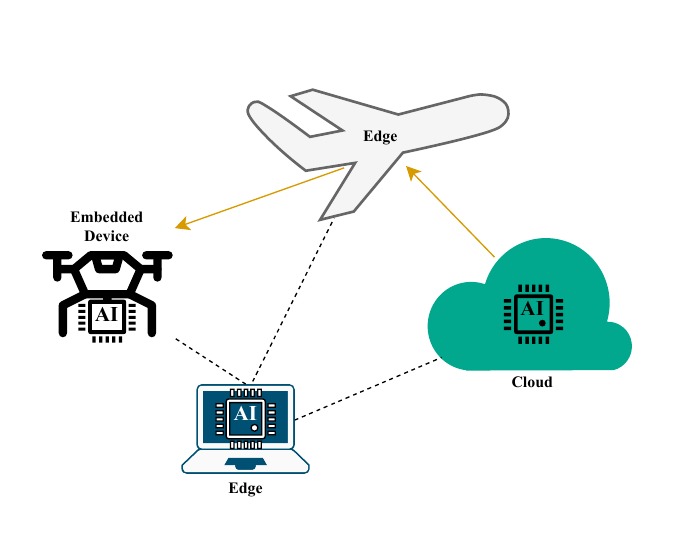}
         \caption{LVLMs can be deployed on the device (drones), at the edge (ground control stations, other aerial vehicles), or in the cloud. }
         \label{fig:edge}
\end{figure}

\noindent \textbf{LVLMs and Drones:} An additional consideration in our problem formulation is where LVLM-based reasoning can realistically be executed. One option is to run small models directly \emph{onboard} the drone, which minimizes latency and dependence on communication links, but requires lightweight models that trade reasoning depth for efficiency. A second option is to leverage \emph{edge devices} such as ground stations, nearby base stations, or, in the case of swarms, a more powerful UAV, which provides more compute capacity than onboard processors while still offering relatively low latency communication; this setting enables medium-sized LVLMs that balance accuracy and resource demands. Finally, \emph{cloud deployment} would require our drone to talk to the cloud, allowing us to use the largest and most capable models, but introducing additional latency, network dependence, and potential reliability issues if connectivity is lost during an emergency. These options are illustrated in Figure~\ref{fig:edge}.

In practice, the choice of deployment setting reflects a tradeoff between the hardware available to the platform and the size and complexity of the LVLM that can be used for real-time decision making. Our focus in this chapter is on the algorithmic pipeline and its reasoning capabilities, but the deployment tier—onboard, edge, or cloud—will ultimately determine the operational envelope for embodied AI in safety-critical drones. In our experimental section, we will use three different-sized models, representing the capabilities of onboard, edge, or cloud LVLMs; in particular, our implementation will consider GPT-5 Nano, GPT-5 Mini, and GPT-5. For practical uses, however, the embedded or edge LVLM will likely be an open model, such as InternVL~\cite{chen2024internvl}, as black-box LVLMs accessible only via cloud APIs cannot be deployed directly on drones.

\section{Design Tradeoffs and Pipeline Overview}

Designing a recovery pipeline around LVLMs forced us to confront the tension between adaptability and reliability. We knew that a purely end-to-end approach—letting the LVLM decide directly from raw images where and how to land—would be maximally flexible, but too risky for safety-critical decisions: hallucinations, inconsistent outputs, or latency spikes could easily cascade into catastrophic failures. At the other extreme, a fully hand-engineered system would be predictable but brittle, unable to cope with novel or unexpected events. Our solution was to adopt a hybrid, modular design: give the LVLM responsibility only for the semantic judgments where its common-sense reasoning is indispensable, and anchor those judgments with conventional perception and control methods that are faster and more predictable.

\begin{figure}[t]
    \centering
    \includegraphics[width=\linewidth]{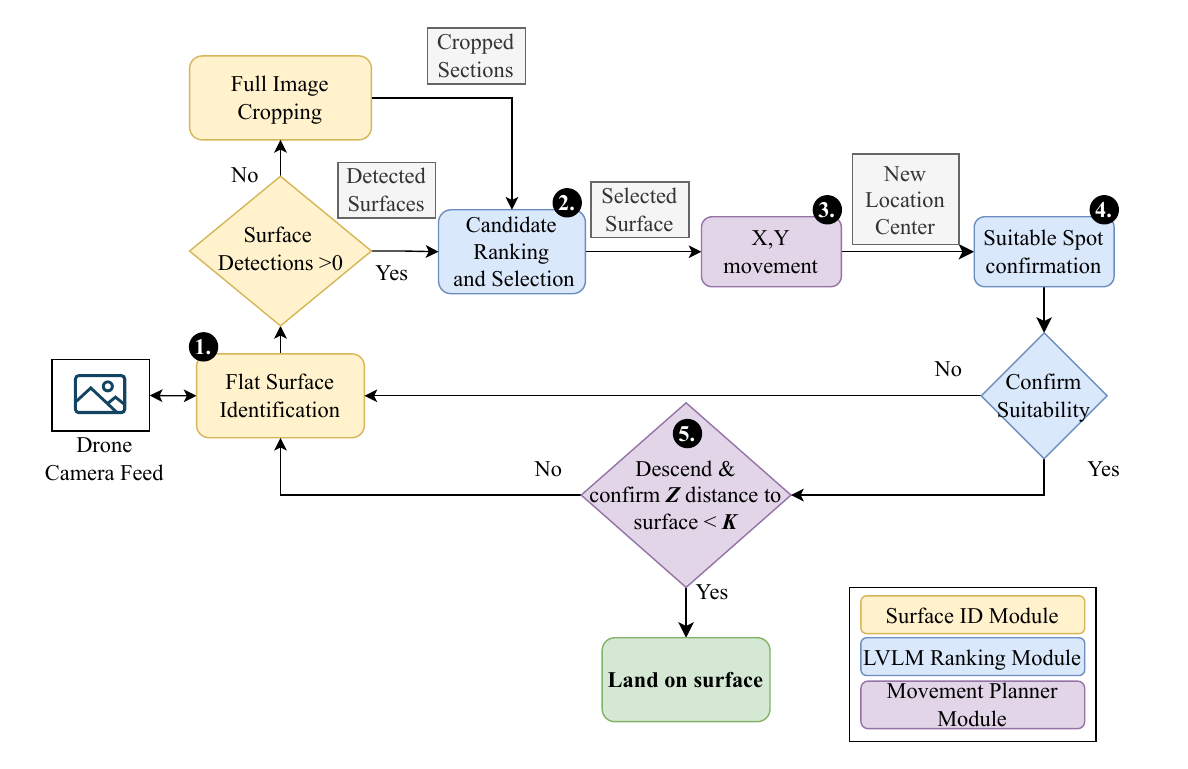}
    \caption{Detailed Pipeline. }
    \label{fig:pipeline}
\end{figure}

This rationale led us to structure the pipeline into three interconnected modules (Figure~\ref{fig:pipeline}). The \textbf{Surface ID Module} handles raw perception and identifies plausible candidate surfaces. The \textbf{LVLM Ranking Module} interprets those candidates, applying contextual reasoning to select and later confirm a safe landing site. Finally, the \textbf{Movement Planner Module} ensures that high-level decisions translate into reliable motion by grounding LVLM outputs in precise world coordinates. This separation of roles keeps the LVLM “in the loop” where it adds unique value, while preventing it from exerting unchecked control over low-level behavior.
\begin{figure}
    \centering
    \includegraphics[width=0.8\linewidth]{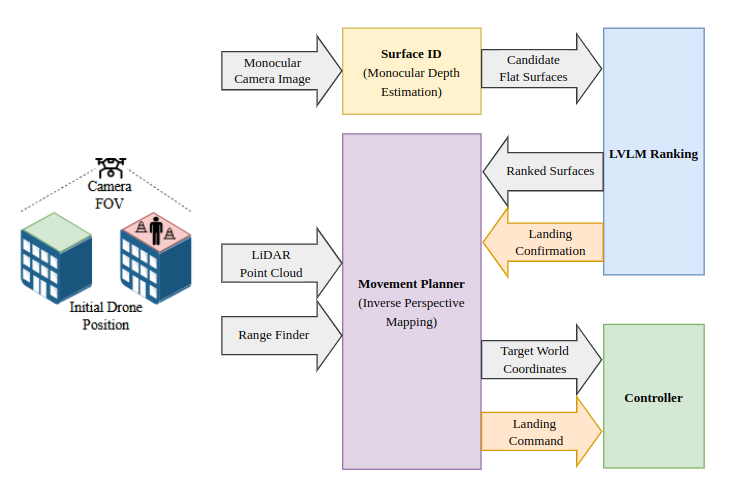}
    \caption{Data flow diagram for the pipeline and its modules. Gray arrows denote transfer of data, orange arrows represent decisions.}
    \label{fig:dataflowdiagram}
\end{figure}
The pipeline uses three sensor modalities: camera images, LiDAR point clouds, and range finder distances. Figure~\ref{fig:dataflowdiagram} shows how data flows between the described modules.

Together, these modules form a loop in which \textbf{(1) candidate surfaces are proposed, (2) semantically evaluated, and then (3) physically reached,} with multiple opportunities for validation before the final descent. The design is deliberately redundant—e.g., the LVLM evaluates surfaces twice, once before motion and once after repositioning—because redundancy is a feature, not a flaw, in safety-critical systems.

\subsection{Surface ID Module} \label{sec: flat_surface_detection}

The first stage of our pipeline is the Surface ID Module, whose task is deliberately narrow: rather than deciding where to land, it reduces uncertainty by pruning the search space and presenting the LVLM with only plausible candidates. This separation keeps the LVLM from being overwhelmed by raw imagery, limits the risk of hallucinations, and ensures that subsequent reasoning focuses on a small number of meaningful options. If no candidates are detected, the module partitions the input image into quadrants so that the LVLM can at least suggest a direction of motion, guaranteeing that the system always produces a next step instead of stalling. The tradeoff here is intentional: we sacrifice some end-to-end flexibility in order to gain robustness, interpretability, and consistent forward progress.

To identify candidate landing zones, we first compute a depth map from the drone’s downward-facing monocular camera. Although a single RGB image does not directly provide depth, modern deep networks can infer it by learning geometric cues such as texture gradients, object sizes, and perspective lines from large training datasets \cite{yang2024depthv2,eigen2014depthmappredictionsingle}. The output is a dense depth map $D(u,v)$, where each pixel coordinate $(u,v)$ is associated with an estimated distance from the camera.

The goal is to detect flat, planar regions in this depth map. Intuitively, a flat surface (like a rooftop or a road segment) will show a consistent change in depth across neighboring pixels, whereas cluttered or sloped areas will produce irregular variations.

We use the identified flat pixel regions to create a mask $M(u,v)$, which we refine through morphological filtering to remove noise from textures or depth artifacts, yielding a cleaned mask $M'(u,v)$. Connected components are then extracted, and any connected component that exceeds a minimum area threshold is marked as a valid flat surface. Finally, for each valid region, we compute a bounding box that localizes the surface within the original RGB image. These cropped patches are passed to the LVLM Ranking Module for semantic evaluation.

\subsection{LVLM Ranking Module.} 

The LVLM then evaluates the candidate surfaces with prompts designed to enforce safety and reduce hallucinations. We deliberately place the LVLM here—between perception and control—because its ability to interpret context is most useful at this semantic level. To address the challenge of dynamic environments, we invoke the module twice: once to rank the candidates before movement, and again to confirm the safety of the chosen site when the drone has repositioned. This double check introduces latency, but provides robustness by ensuring that conditions have not changed while moving.

In practice, the LVLM interaction is separated into two stages. In the \textbf{ranking stage}, the model receives up to five cropped candidate images from the Surface ID Module and orders them by suitability, considering both flatness and the presence of hazards. The top-ranked region is then passed to the Movement Planner. Once we reach the intended destination, we enter the \textbf{confirmation stage}, where the LVLM is queried again with a close-up view to verify that the site is still safe for descent. This two-pass process provides the system with an up to date evaluation of the target, to address the challenge of surfaces whose safety may change dynamically over time.

\begin{figure*}[ht]
    \centering  
    \includegraphics[width=\textwidth]{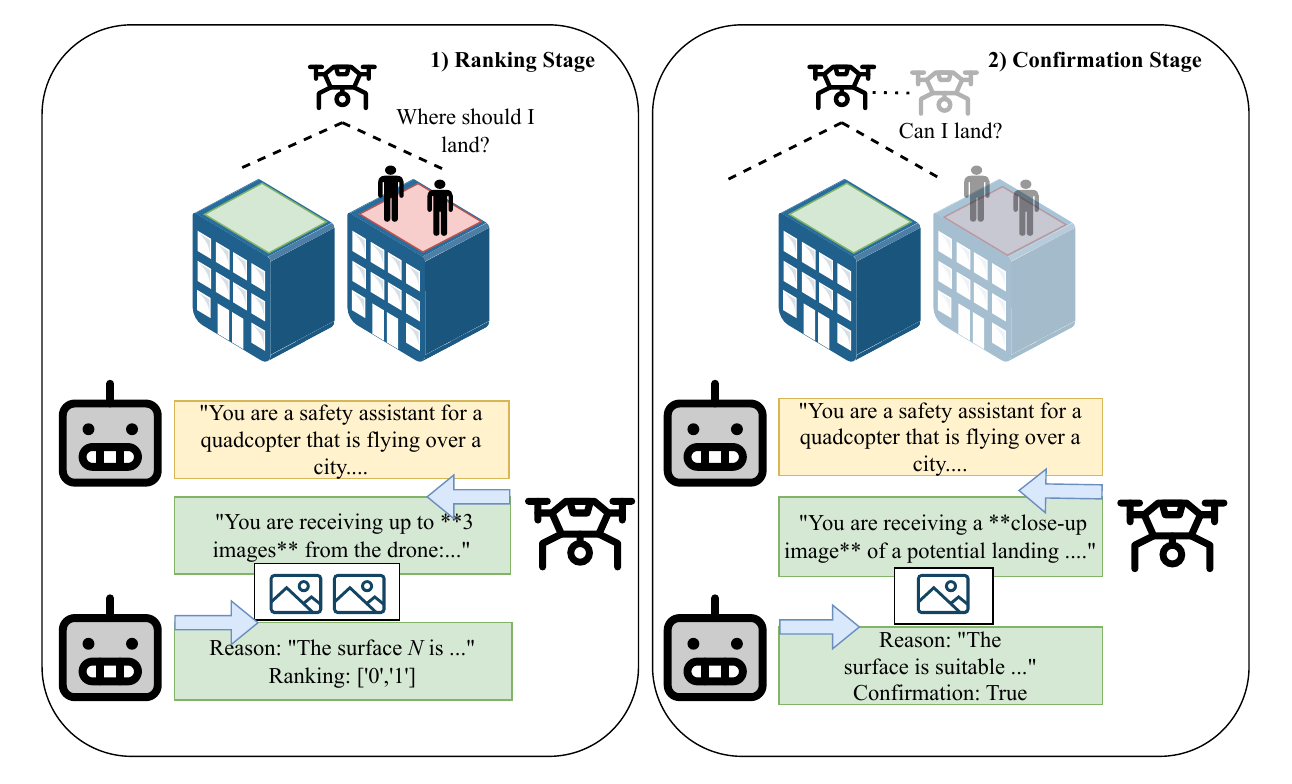}
    \caption{Two-stage conversational prompting. }
    \label{fig:prompts}
\end{figure*}

To maximize reliability, we carefully design prompts for both stages. A system prompt frames the LVLM as an emergency landing assistant, specifying its role, expected inputs, and output format. Conversational prompts (Fig.~\ref{fig:prompts}) are structured using a Chain-of-Thought approach~\cite{wei2023chainofthoughtpromptingelicitsreasoning,sahoo2025systematicsurveypromptengineering}, encouraging the model to reason step by step rather than guessing directly. Safety constraints are reinforced through imperative formulations such as “MUST”~\cite{amatriain2024promptdesignengineeringintroduction}, which obligate the LVLM to apply strict selection rules and always return a valid choice. While the prompts for ranking and confirmation share this structure, their emphasis differ: the ranking stage focuses on general surface suitability, while the confirmation stage focuses narrowly on detecting immediate threats that could compromise the final descent. The details of the prompts are in the Appendix.

A final threshold check on the vertical distance $Z$ ensures that the drone is close enough for a conventional landing controller to take over, bridging LVLM-based reasoning with the reliability of low-level control.

\subsection{Movement Planner Module.} 

The LVLM’s decision is expressed as an image crop, and we take the center of this crop as the target destination. To act on this decision, we must convert that central pixel in the camera image into a physical 3D location that the drone can navigate to. This translation from image space to world space is the purpose of inverse perspective mapping (IPM).

The process unfolds in three steps:

\textbf{1. From LiDAR to image space.} We first build a mapping between LiDAR points and image pixels using the pinhole camera model. A LiDAR point $p=(x_{frd},y_{frd},z_{frd})$ in the drone’s body frame (Forward–Right–Down) projects into the image plane as
\begin{equation}
u = \frac{x_{frd}\cdot f}{z_{frd}} + c_x, \quad
v = \frac{y_{frd}\cdot f}{z_{frd}} + c_y,
\end{equation}
where $(u,v)$ are image coordinates, $f$ is the focal length, and $(c_x,c_y)$ is the camera’s principal point. Repeating this for all LiDAR points creates a lookup table linking pixels to 3D voxels.

\textbf{2. From image to 3D.} When the LVLM outputs a candidate image, we take its center pixel $(u,v)$ and look up the corresponding LiDAR point. If the exact pixel has no LiDAR match (due to sparsity), we select the nearest pixel with a valid correspondence. The result is a 3D point $(x_{frd},y_{frd},z_{frd})$ in the drone’s body frame.

\textbf{3. From body to world.} To express this location in the world frame, we rotate the FRD coordinates into the North–East–Down (NED) frame using the drone’s yaw angle $\psi$:
\begin{equation}
    \begin{bmatrix}
        x_{ned} \\
        y_{ned} 
    \end{bmatrix}
    =
    \begin{bmatrix}
        \cos\psi & -\sin\psi \\
        \sin\psi &  \cos\psi
    \end{bmatrix}
    \begin{bmatrix}
        x_{frd} \\
        y_{frd}
    \end{bmatrix}.
\end{equation}
We then provide $(x_{ned},y_{ned})$ to the controller, which moves the drone horizontally above the chosen landing site. The vertical component is ignored at this stage; descent begins only after safety has been reconfirmed.

Once the drone has moved above the candidate site, the LVLM Ranking Module is invoked again for confirmation. If the site is still deemed safe, the drone uses a distance sensor to measure altitude. It first descends to a fraction $k$ of this distance, giving the system another opportunity to reevaluate safety if new hazards appear. If the drone is already within a 5-meter threshold, it proceeds directly to final landing.

This design keeps the LVLM responsible only for high-level judgments—identifying which region looks safe—while the Movement Planner handles the geometric and control details needed to reach that region. In this way, the LVLM “suggests” and traditional control “guarantees,” providing layered defenses against unsafe outcomes.

\section{Implementation}

\noindent{\textbf{Environment.}} To evaluate our pipeline, we build on the Unreal Engine 5 City Sample Project, a large-scale, photorealistic urban environment originally designed to showcase next-generation rendering. The City Sample is uniquely well suited to testing embodied AI for drones: it combines realistic rooftops, roads, vehicles, and pedestrians with full configurability, including weather and lighting. Unlike many static simulators, its scenarios are dynamic, with cars and pedestrians moving through the streets, creating the kinds of evolving conditions that challenge real-world decision-making. This realism is also practically valuable: recent work has demonstrated that synthetic data generated from the City Sample improves object detection models, thanks to its high fidelity and dynamic variety~\cite{turkcan2024boundless}.

To make scenarios more challenging, we added obstacles such as HVAC units to otherwise safe rooftops. The high degree of customization in City Sample allows future extensions. In the future, we plan to further stress-test our pipeline with rooftop fires, obstructing pedestrians, or adversarial markings (e.g., text or symbols) to probe system robustness under more challenging scenarios.

\noindent{\textbf{Drone.}} We integrated our pipeline with Unreal Engine~5 using Cosys-AirSim, a plugin developed by Cosys-Lab as an extension of Microsoft’s AirSim. Cosys-AirSim lets us simulate drones with realistic physics and visuals, and it provides APIs that allow our pipeline to stream sensor data and issue motion commands in real time. It also handles the conversion between Unreal’s coordinate system and the North–East–Down (NED) convention used in robotics.

During our experiments, we ran the built-in \texttt{SimpleFlight} controller. We configured the simulated drone with the sensors required for our pipeline: a downward-facing RGB camera for perception and LVLM queries, a distance sensor for controlled descent, a LiDAR sensor for inverse perspective mapping, and IMU/GPS data for pose reference.

\noindent{\textbf{Surface Detection.}} We identify candidate landing surfaces using monocular depth estimation. Specifically, we employ a compact version of the Depth Anything V2 model \cite{yang2024depthv2}, which predicts dense depth maps from single RGB images captured by the drone’s downward-facing camera. From these maps, we segment locally flat regions by thresholding gradient magnitudes and applying simple morphological filtering to remove noise. We then extract connected components that exceed a minimum area threshold and generate bounding boxes for each valid surface. These cropped image patches serve as candidate landing zones and are passed to the LVLM for semantic evaluation and ranking.

\noindent{\textbf{LVLM.}} We evaluated cloud deployed models, three OpenAI multimodal models of different scales—GPT-5, GPT-5-mini, and GPT-5-nano alongside Gemini Flash 2.5, Mistral 3 Large and a model deployed on an edge device, InternVL 2.5, in order to study the tradeoff between reasoning strength and computational efficiency. GPT-5 and Gemini Flash 2.5 provide the most powerful multimodal reasoning and serve as an upper bound for performance. GPT-5-mini delivers nearly comparable accuracy while requiring fewer resources, making it attractive for edge deployment. GPT-5-nano is the most compact and recent GPT model, and represents state of the art models that could be run on constrained hardware. Likewise, Mistral 3 and InternVL are small open-source models that could potentially run on an edge device and have faster response time.  This range of models allows us to explore how the size and capacity of an LVLM affect decision-making in safety-critical drone scenarios.

\section{Evaluation Metrics}

To evaluate the pipeline, we first tested each module in isolation (20 iterations per module) before moving to end-to-end experiments. This modular evaluation allowed us to verify that every component operated as intended and to quantify its specific contribution to the overall system. We relied on three metrics to assess module performance.
  
\noindent \textbf{Jaccard Index (JI)}: The Jaccard Index (JI) \cite{10.1007/s11263-009-0275-4} quantifies how well a predicted bounding box overlaps with the ground-truth surface. A value of 1 indicates perfect agreement. For a ground-truth area $B$ and a predicted area $A$, we define
\begin{equation}
    JI = \frac{A \cap B}{A \cup B}.
    \label{eq:iou}
\end{equation}
We use this metric to evaluate the accuracy of the Surface ID module.

\noindent \textbf{Landing Distance.} We evaluate the precision of the Movement Planner by measuring the Euclidean distance between the drone’s final landing position and the ground-truth target in the UE5 coordinate space. Let $(x_{gt}, y_{gt})$ denote the ground-truth location and $(x_r, y_r)$ the position reached by the drone. The distance is
\begin{equation}
   d = \sqrt{(x_{gt} - x_r)^2 + (y_{gt} - y_r)^2}.
   \label{eq:euclidean}
\end{equation}
A smaller value indicates more accurate guidance toward the intended landing site.

\noindent \textbf{LVLM Success Rate.} To evaluate the LVLM, we designed two test scenarios, each with a pair of candidate rooftops: one clear of obstacles and one cluttered with debris such as rubble or HVAC units. In each trial, success is counted when the LVLM selects the clear rooftop. The success rate is the percentage of trials in which the model made the correct choice.

\section{Benchmarking Modules and the Pipeline in Curated Scenarios}

\begin{figure*}[ht]
    \centering  
    \hfill\begin{subfigure}[b]{0.49\textwidth}
         \includegraphics[width=\textwidth]{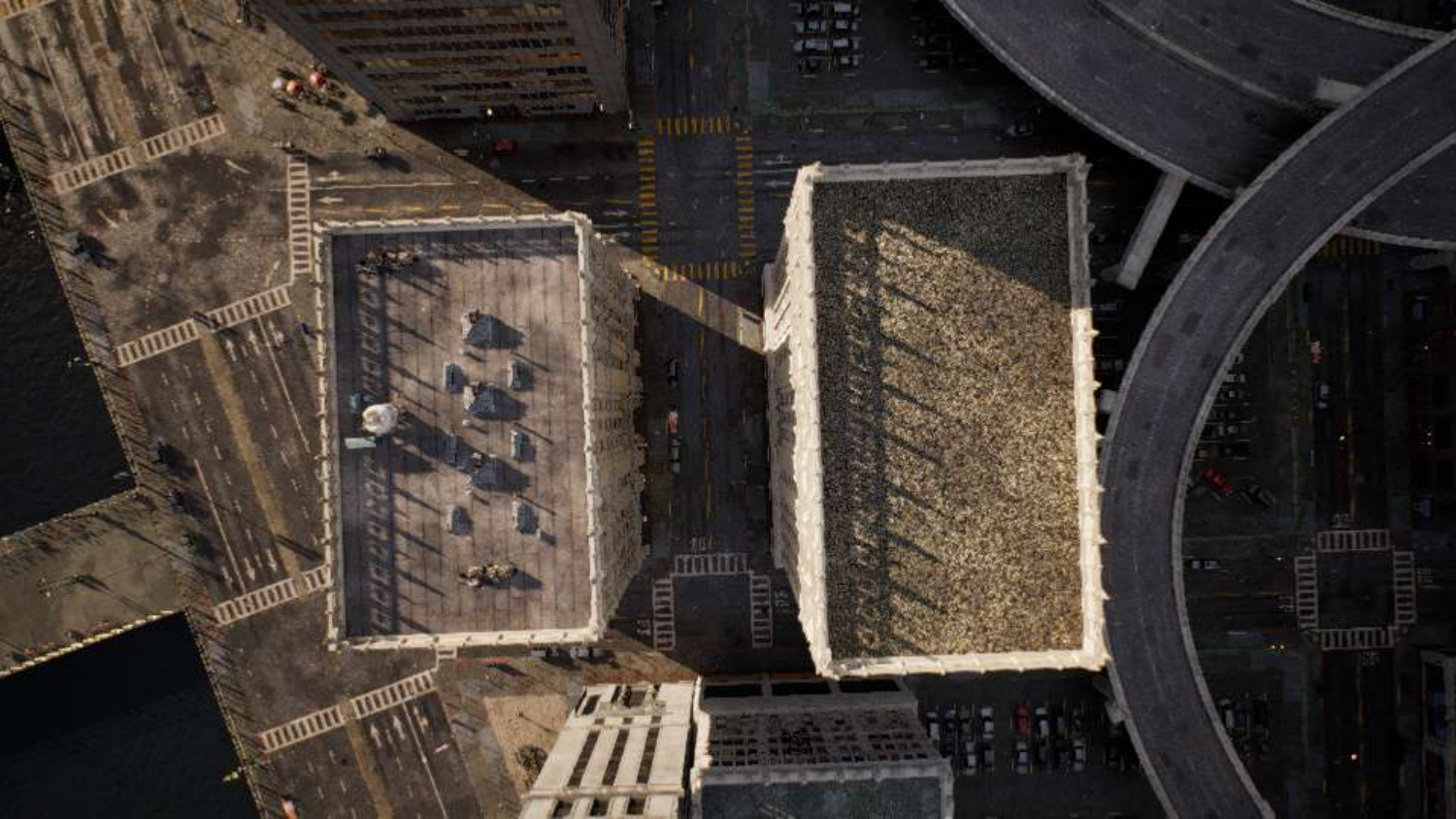}
         \caption{Scenario 1. }
    \end{subfigure}
    \hfill
    \begin{subfigure}[b]{0.49\textwidth}
         \includegraphics[width=\textwidth]{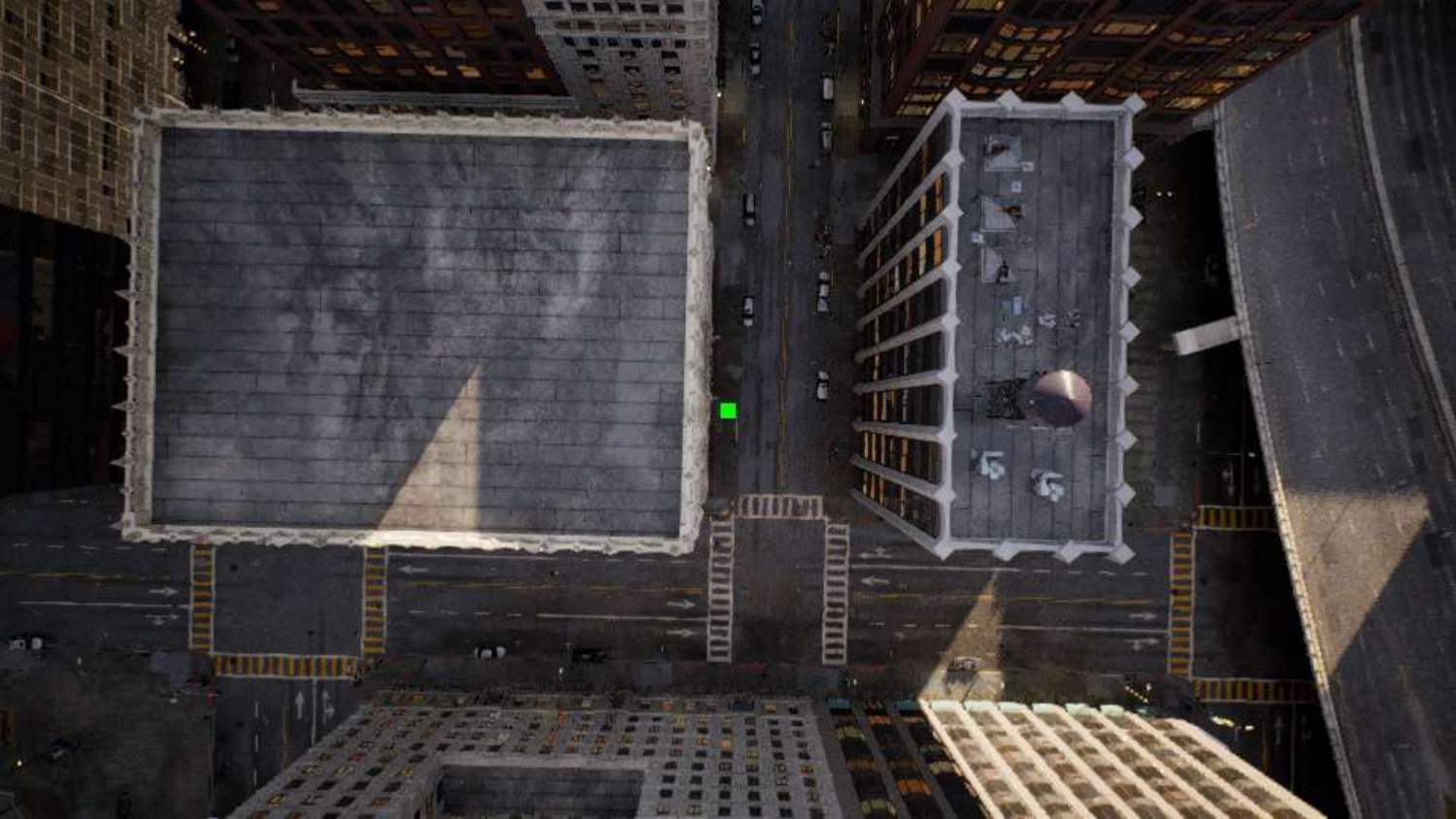}
         \caption{Scenario 2.}
    \end{subfigure}
    \caption{Testing Scenarios, Original views} 
    \label{fig:scenarios}
\end{figure*}

For our initial tests we use two fixed scenarios, as illustrated in Fig. \ref{fig:scenarios}. Both scenarios have two rooftops, one cluttered with obstacles and one clear of any obstacles (this should be the preferred landing option).

\subsection{Results of Individual Modules}

Table~\ref{tab:successRateJIDist} summarizes the performance of the Surface ID and Movement Planner modules. The Surface ID Module consistently detected the correct candidate surface, with a higher Jaccard Index in Scenario~1 (87\%) and a lower but still acceptable score in Scenario~2 (60\%). After selecting the center-point of these locations, the Movement Planner guided the drone to within roughly one meter of the ground-truth target in both scenarios, with slightly larger drift in Scenario~2. These results show that both modules provide accurate and reliable inputs for the overall pipeline.

\begin{table}[h!]
\centering
\caption{Success Rate for Surface ID and Movement Planner modules}
\label{tab:successRateJIDist}
\begin{tabular}{lcc}
\toprule
 & JI Score Avg. & Avg. Distance (m)\\
\midrule
  Scenario 1   & 87.09\% & 0.802\\
 Scenario 2  & 59.55\%     & 1.145 \\
\bottomrule
\end{tabular}
\end{table}

 




\begin{table}[h!]
\centering
\caption{Operation Success Rate in Scenario 1 and Scenario 2 for LVLM Operations}
\label{tab:successRateScenario2}
\begin{tabular}{llcccccc}
\toprule
 & Stage 
 & GPT-5
 & \begin{tabular}[c]{@{}c@{}}GPT-5\\mini\end{tabular}
 & \begin{tabular}[c]{@{}c@{}}GPT-5\\nano\end{tabular}
 & \begin{tabular}[c]{@{}c@{}}Gemini\\2.5 F\end{tabular}
 & \begin{tabular}[c]{@{}c@{}}Mistral\\3 L\end{tabular}
 & \begin{tabular}[c]{@{}c@{}}InternVL\\2.5\end{tabular} \\ 
\cmidrule(l){2-8}

\multirow{2}{*}{Scenario 1} 
 & Rank.        & 100\% & 100\% & 20\%  & 100\% & 100\% & 100\% \\ 
 & Conf.   & 100\% & 95\%  & 100\% & 100\% & 90\%  & 100\% \\ 

\multirow{2}{*}{Scenario 2} 
 & Rank.       & 100\% & 100\% & 100\% & 100\% & 100\% & 100\% \\ 
 & Conf.   & 100\% & 75\%  & 20\%  & 100\% & 20\% & 100\% \\ 
\bottomrule
\end{tabular}
\end{table}

Table~\ref{tab:successRateScenario2} compares the performance of different LVLMs in the Ranking Module. Most model such as Gemini 2.5 F., GPT-5 and GPT-5-mini consistently selected the correct landing surface in both scenarios. However, differences start to appear on the confirmation step with GPT-5-mini and Mistral showing slightly lower accuracy on the confirmation step in Scenario~1, likely due to a more cautious bias, but overall demonstrated the potential of smaller models.

In contrast, GPT-5-nano performed much worse. It struggled with ranking in Scenario~1 (20\%) and failed the confirmation step in Scenario~2 (20\%), often rejecting valid rooftops by labeling tiles as ``shingled,'' ``irregular,'' or ``uneven.'' Similarly, Mistral 3 L. starts showing its limitations by failing most of the iterations in the Scenario~2 confirmation step (20\%), repeatedly mistaken the roofs as ''brick'' or ''stone'' walls.   



\begin{figure*}[ht]
    \centering
    \begin{subfigure}[b]{0.31\textwidth}
        \includegraphics[width=\textwidth]{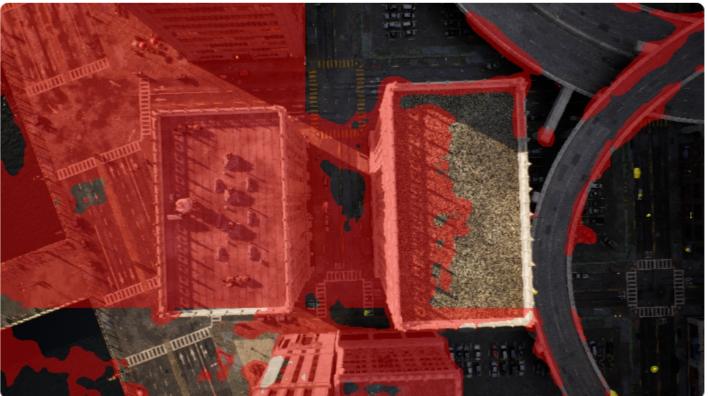}
        \caption{OpenLander \\Segmentation}
    \end{subfigure}
    \hfill 
    \begin{subfigure}[b]{0.31\textwidth}
        \includegraphics[width=\textwidth]{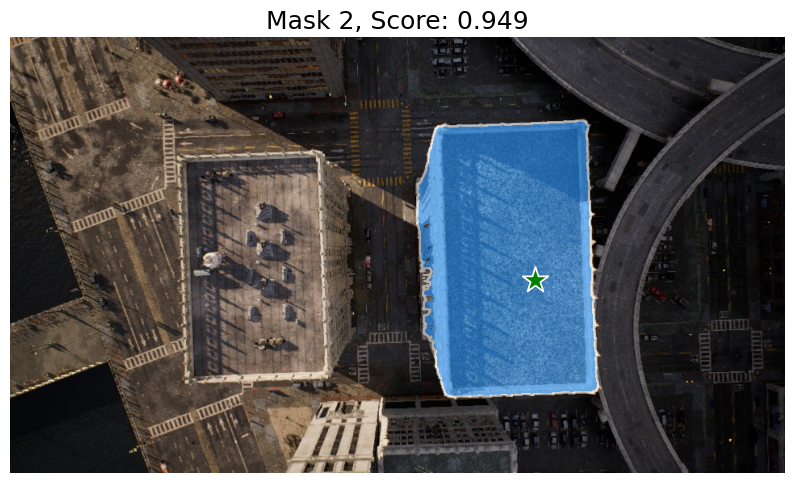}
        \caption{SAM Point \\Segmentation}
    \end{subfigure}
    \hfill
    \begin{subfigure}[b]{0.31\textwidth}
        \includegraphics[width=\textwidth]{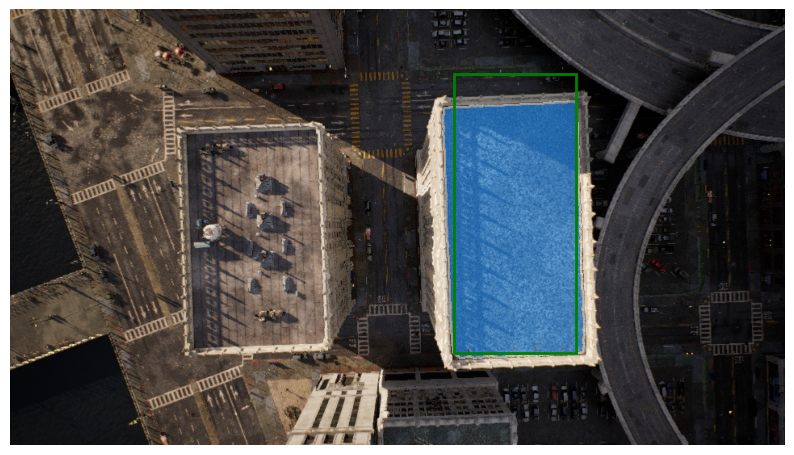}
        \caption{SAM Bounding \\ Box}
    \end{subfigure}
    \caption{Comparison of vision-based approaches for Scenario 1.}
    \label{fig:vision_comparison}
\end{figure*}

\subsection{Qualitative comparison with Vision-Based Baselines}
To further contextualize our results, we performed a qualitative and quantitative comparison using two established vision-based baselines on our ground-truth data: OpenLander \cite{OpenLander} and the Segment Anything Model (SAM) \cite{SAM}. 

As illustrated in Figure 1.6, SAM demonstrates exceptional geometric precision. When evaluated against our ground-truth landing coordinates using Intersection over Union (IoU), SAM achieved a score of 0.951 in Scenario 1 and 0.787 in Scenario 2. These scores indicate that SAM can isolate potential landing surfaces with high spatial accuracy. 

However, this geometric proficiency comes with significant trade-offs. First, SAM is purely objective; it provides high-fidelity segmentation but lacks the semantic capacity to distinguish between a safe landing zone and a restricted or dangerous area, such as an active highway or a private balcony. Second, while we could use SAM to refine our candidate proposals, its computational overhead is substantial. Given the hardware constraints of small UAVs, running a large model like SAM alongside our depth map estimation module would likely increase latency and potentially compromise the drone's ability to respond to sudden events.

Similarly, OpenLander’s safety masks successfully identify "flatness" and provide basic risk-mitigation categories, such as obstacles or humans. Yet, as seen in our urban evaluations, it fails to incorporate the higher-level common-sense reasoning required to navigate evolving urban contexts, such as predicting the reappearance of traffic on a currently empty roadway, a capability that our LLM-driven pipeline addresses through contextual reasoning.

 \subsection{LVLM Context Differences}

 \begin{figure*}[ht]
    \begin{subfigure}[b]{0.28\textwidth}
         \includegraphics[width=\textwidth]{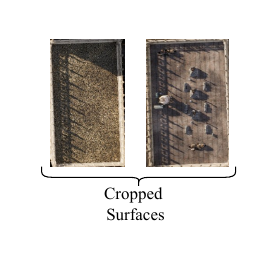}
         \caption{Cropped View}
    \end{subfigure}
    \centering 
    \begin{subfigure}[b]{0.71\textwidth}
         \includegraphics[width=\textwidth]{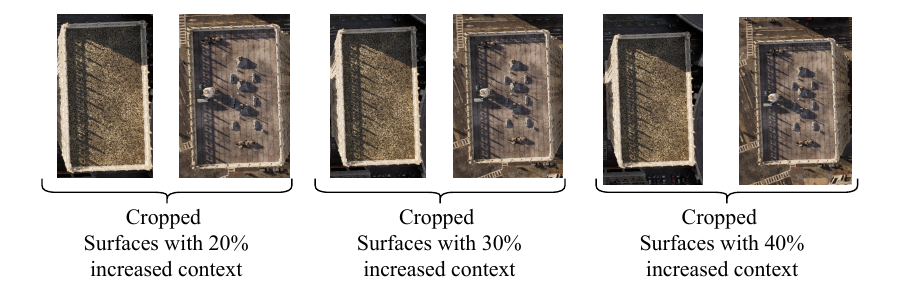}
         \caption{Increased Partial Context Views}
    \end{subfigure}    
    \centering 
    \begin{subfigure}[b]{0.69\textwidth}
         \includegraphics[width=\textwidth]{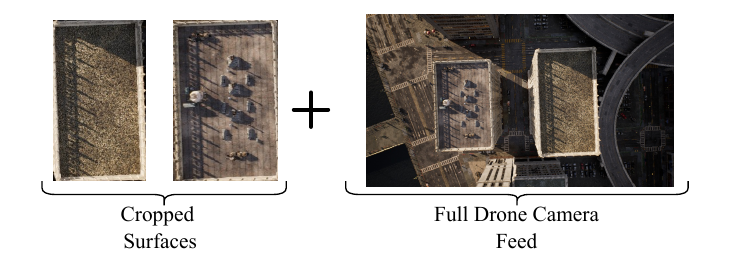}
         \caption{Full Image View}
    \end{subfigure}
    \caption{Different Context Levels Provided to the LVLM}
    \label{fig:cont_diffs}
\end{figure*}

\begin{figure*}[ht]
    \centering  
    
    \begin{subfigure}[b]{0.6\textwidth}
         \includegraphics[width=\textwidth]{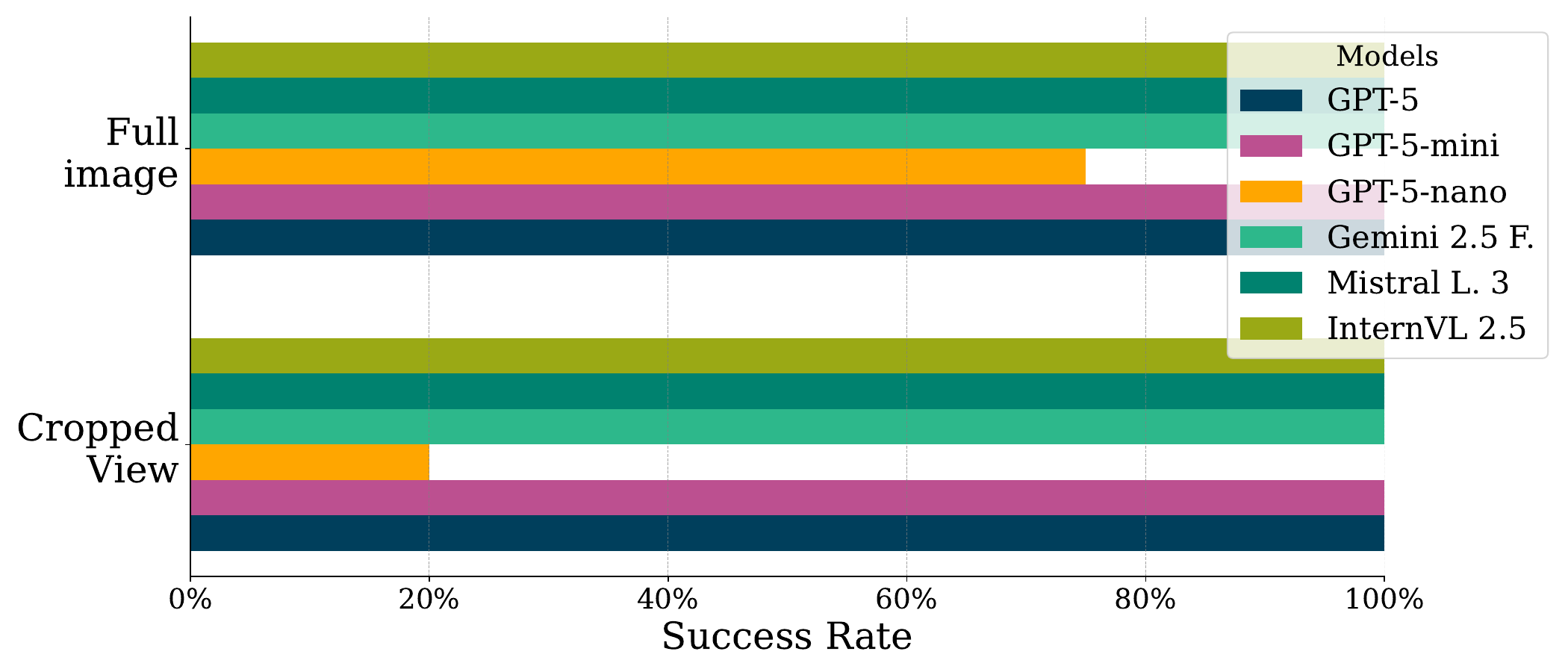}
         \caption{Scenario 1}
    \end{subfigure}
    \vspace{1em}
    \centering 
    \begin{subfigure}[b]{0.6\textwidth}
         \includegraphics[width=\textwidth]{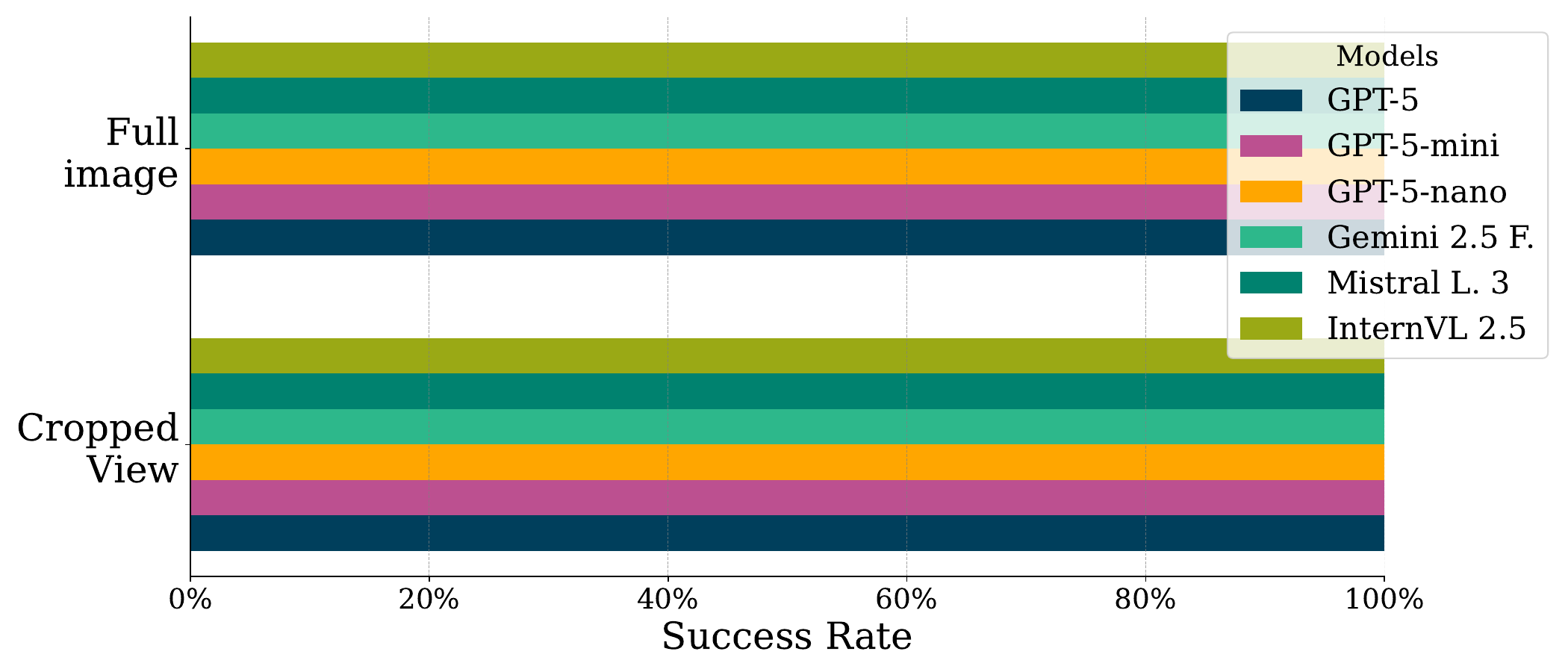}
         \caption{Scenario 2}
    \end{subfigure}
    \caption{Full image view and cropped view evaluation.} 
    \label{fig:ind_context}
\end{figure*}

To probe how input context affects LVLM decisions, we varied the information provided to each model (Fig.~\ref{fig:cont_diffs}). In addition to cropped candidate surfaces from the Surface ID Module, we tested two variations: (i) adding 20--40\% padding around the cropped region, and (ii) supplying the entire camera image alongside the cropped surface.

Figure~\ref{fig:ind_context} and Table~\ref{tab:paddings} summarize the results. Most models were robust across all conditions, achieving perfect accuracy regardless of context in Scenario~2 and consistently correct choices in Scenario~1. In contrast, GPT-5-nano showed pronounced sensitivity. With partial padding, its accuracy fluctuated between 15–25\%, and at times it misclassified safe rooftops as unsuitable due to surface textures such as gravel or shingles. Providing the full image greatly improved Nano’s performance (up to 70\%), yet it still fell short of the larger models. Similarly, InternVL struggled only with Scenario 1 being unable to select the correct surface when context of 40\% margin was added, consistently describing the surface as ''covered in a dense layer of objects'' while stating that the incorrect surface had less objects making it more suitable to land. 

These findings indicate that while larger LVLMs reliably focus on relevant features even with minimal input, most lightweight models benefit substantially from additional context. Supplying the full scene helps smaller models better situate candidate surfaces in their surroundings, reducing—but not eliminating—their cautious bias.




\begin{table}[h!]
\centering
\caption{Success per Context Level in Scenario 1}
\label{tab:paddings}
\resizebox{\linewidth}{!}{%
\begin{tabular}{lcccccc}
\toprule
Partial Context\\ Level & GPT-5 & GPT-5 mini & GPT-5 nano & Gemini 2.5 F. & Mistral L.3 & InternVL 2.5 \\ 
\midrule
20\% & 100\% & 100\% & 15\% & 100\% & 100\% & 100\% \\ 
30\% & 100\% & 100\% & 25\% & 100\% & 90\% & 100\% \\ 
40\% & 100\% & 100\% & 15\% & 95\% & 80\% & 0\% \\ 
\bottomrule
\end{tabular}%
}
\end{table}




\subsubsection{Model Latency}

To analyze the feasibility of our approach and how it would differ in different deployment types (cloud, edge). Cloud-deployed models were accessed via the OpenAI API and OpenRouter, while InternVL was deployed on a local device on the same network. We measured latency during the crop-view experiments and calculated the mean across all experiments. We show these results in Table \ref{tab:response_time}.   

Overall, all models demonstrated response times below 30 seconds, indicating they can handle the pipeline's input in a reasonable time. However, the response time of the largest models widely varies, with GPT-5 taking more than six times as long to respond as Gemini 2.5 F. with the same input. This could be due to GPT-5's larger scale and increased computational cost, while Gemini 2.5 F. is optimized for performance. 

Within this context, our edge model, InternVL, achieves lower latency than all GPT models, demonstrating the benefits of reduced network overhead. However, both Mistral 3 L. and Gemini 2.5 F. outperform InternVL, with Gemini 2.5 F. having the lowest latency overall. These results show that although edge models can offer advantages and reduce latency, larger, optimized models can outperform them in both latency and output quality. 

\begin{table}[h!]
\centering
\caption{Average Response Time per Model}
\label{tab:response_time}
\resizebox{\linewidth}{!}{%
\begin{tabular}{lcccccc}
\toprule
 &
GPT-5 &
GPT-5-mini &
GPT-5-nano &
Gemini F. 2.5 &
Mistral L. 3 &
InternVL 2.5 \\
\midrule
\begin{tabular}[c]{@{}l@{}}
Response \\
Time (s)
\end{tabular}  &
25.783 &
15.152 &
16.938 &
3.755 &
7.184 &
8.579 \\
\bottomrule
\end{tabular}%
}
\end{table}



\subsection{Full Pipeline in the Two Scenarios}\label{sec:2scenarios}

\begin{figure*}[ht]
    \centering  
    
    \begin{subfigure}[b]{0.49\textwidth}
         \includegraphics[width=\textwidth]{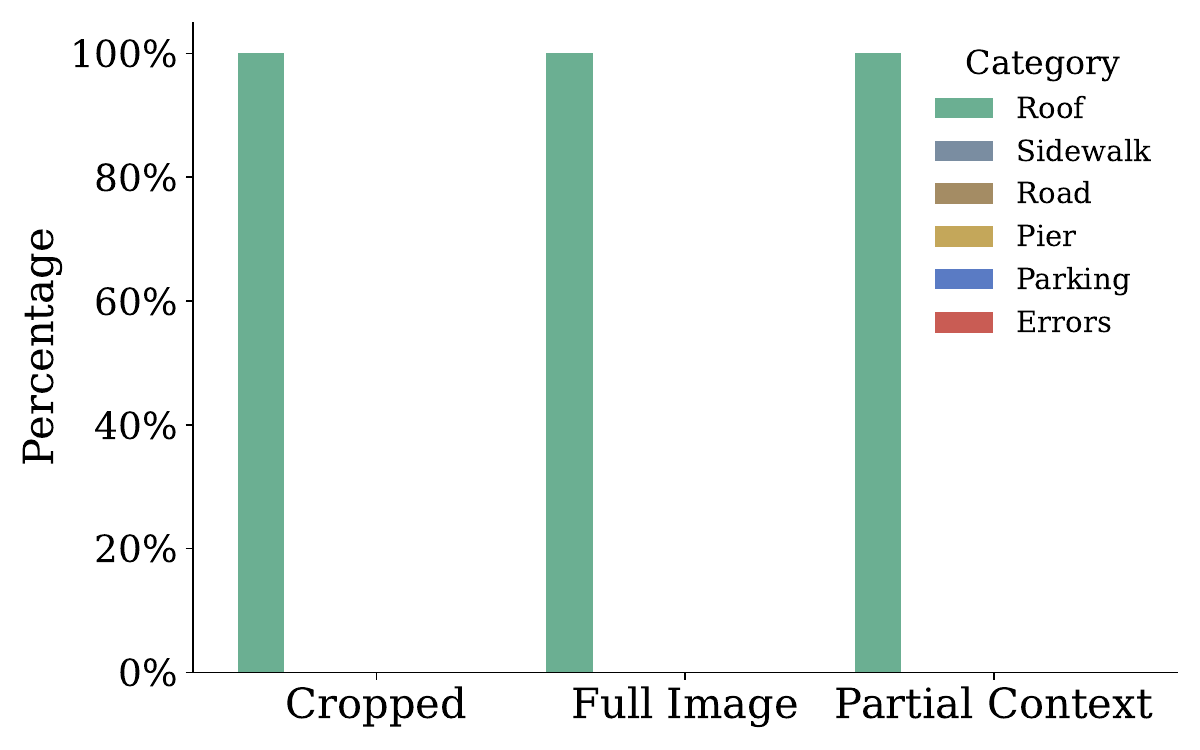}
         \caption{GPT-5}
    \end{subfigure}
    \vspace{1em}
    \centering 
    \begin{subfigure}[b]{0.49\textwidth}
         \includegraphics[width=\textwidth]{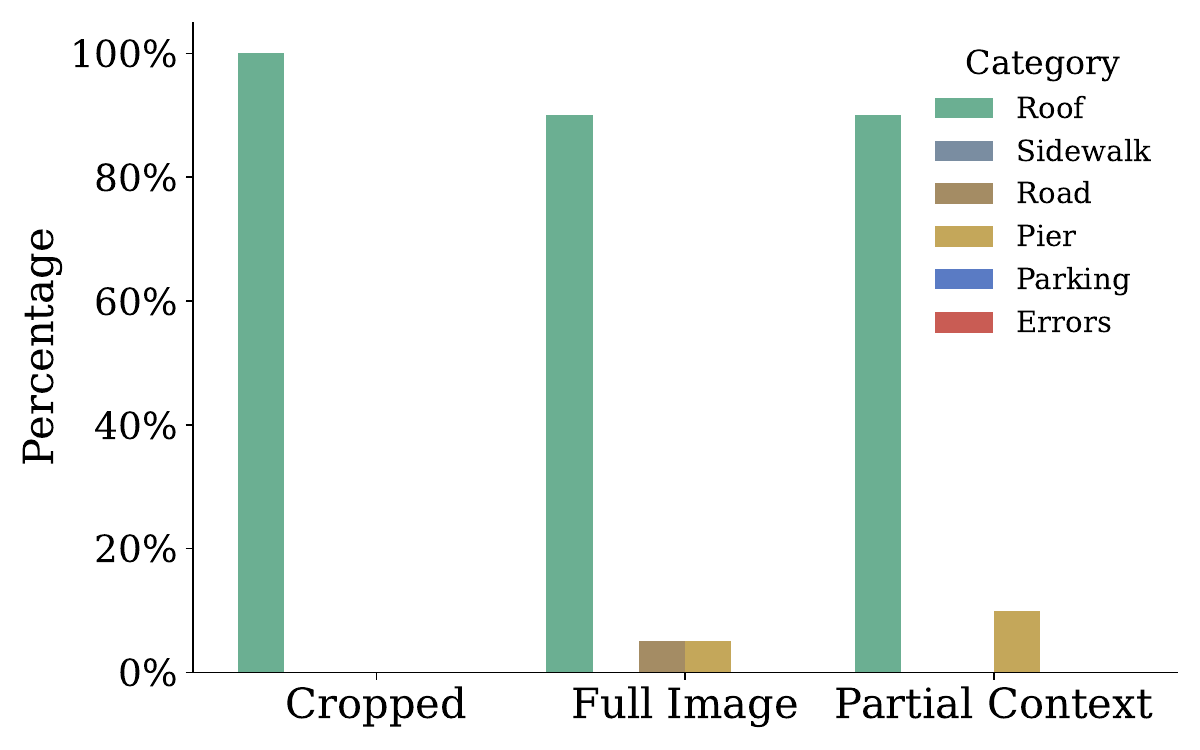}
         \caption{GPT-5 Mini}
    \end{subfigure}    
    \begin{subfigure}[b]{0.49\textwidth}
         \includegraphics[width=\textwidth]{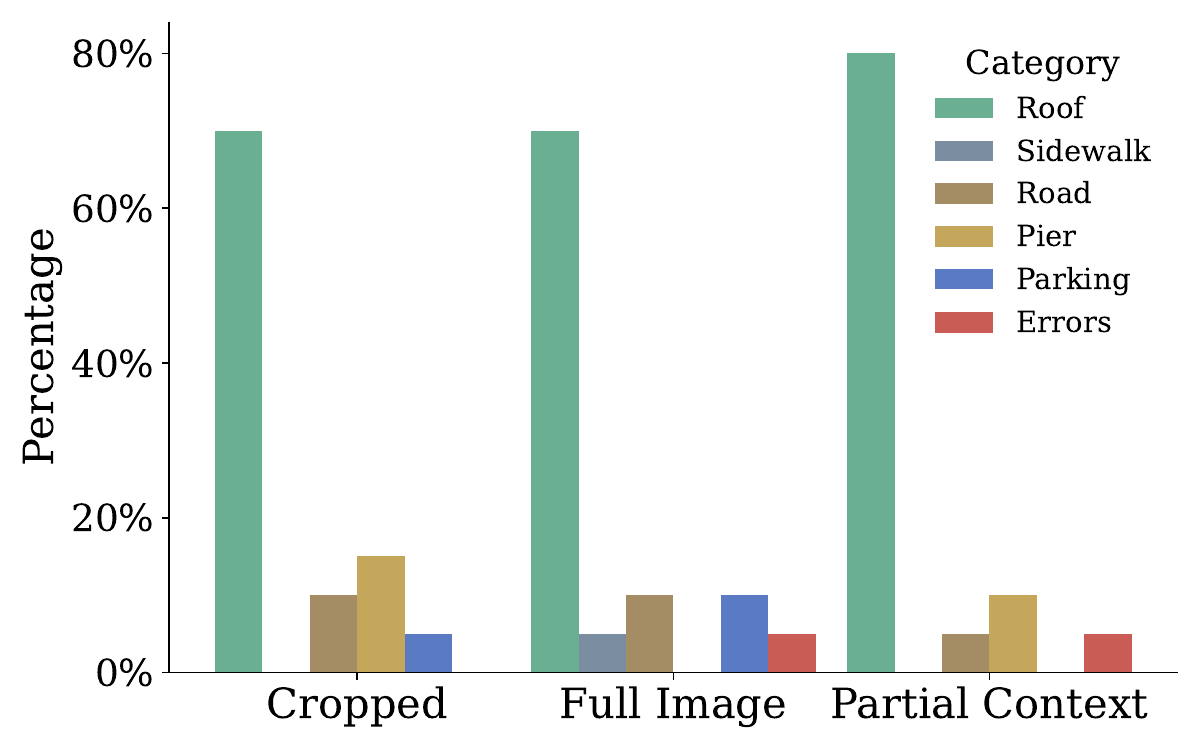}
         \caption{GPT-5 Nano}
    \end{subfigure}
    \begin{subfigure}[b]{0.49\textwidth}
         \includegraphics[width=\textwidth]{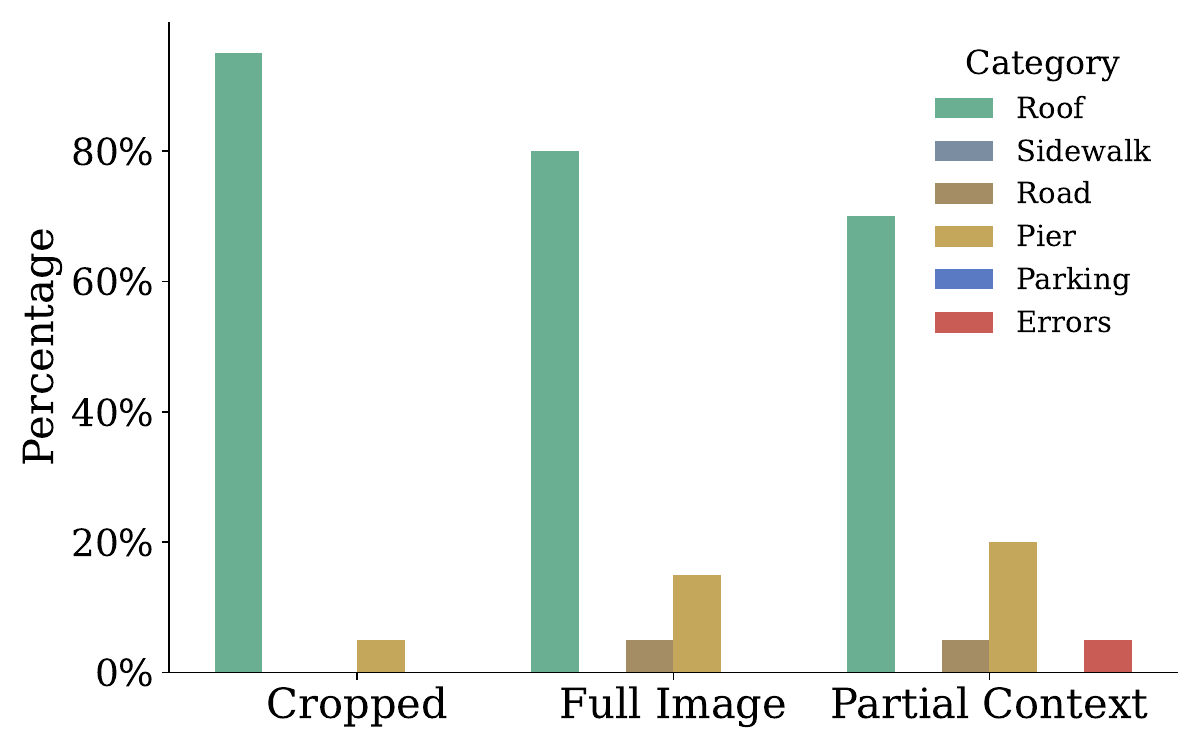}
         \caption{Gemini F 2.5}
    \end{subfigure}
    \begin{subfigure}[b]{0.49\textwidth}
         \includegraphics[width=\textwidth]{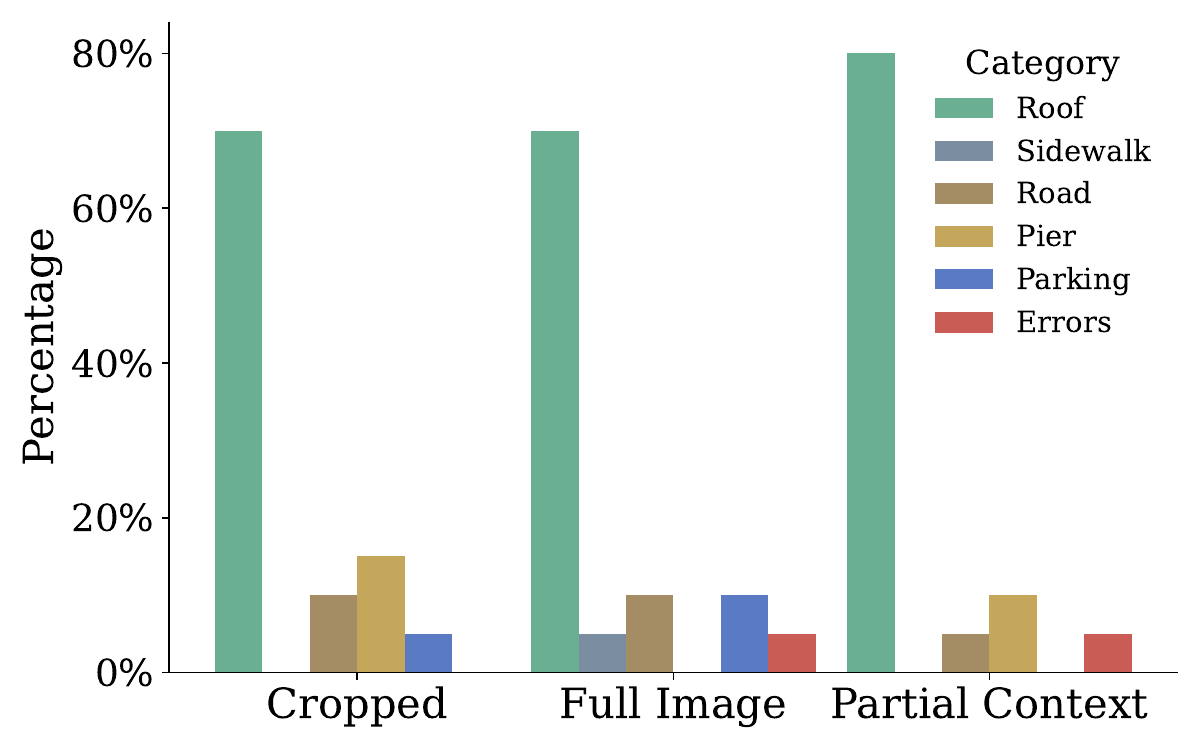}
         \caption{Mistral L 3}
    \end{subfigure}
    \caption{Scenario 1: Type of surface landed on by each model by information level} 
    \label{fig:surfaceType1}
\end{figure*}

\begin{figure*}[ht]
    \centering  
    
    \begin{subfigure}[b]{0.49\textwidth}
         \includegraphics[width=\textwidth]{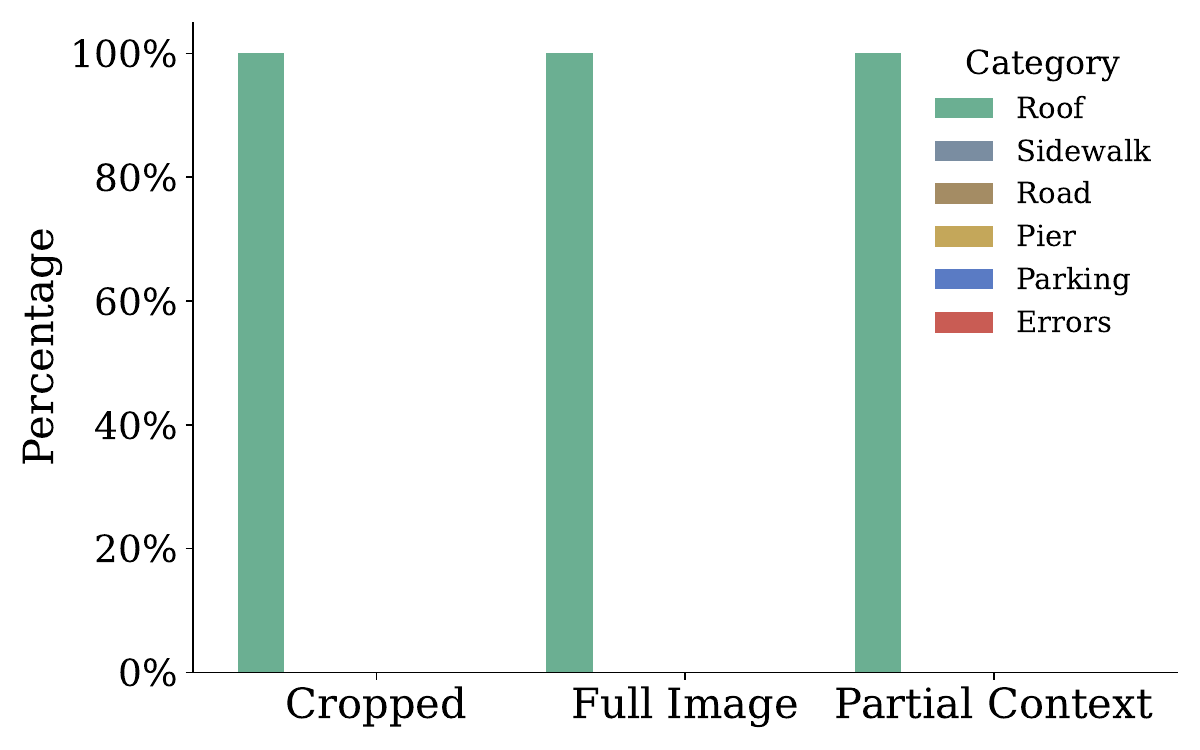}
         \caption{GPT-5}
    \end{subfigure}
    \vspace{1em}
    \centering 
    \begin{subfigure}[b]{0.49\textwidth}
         \includegraphics[width=\textwidth]{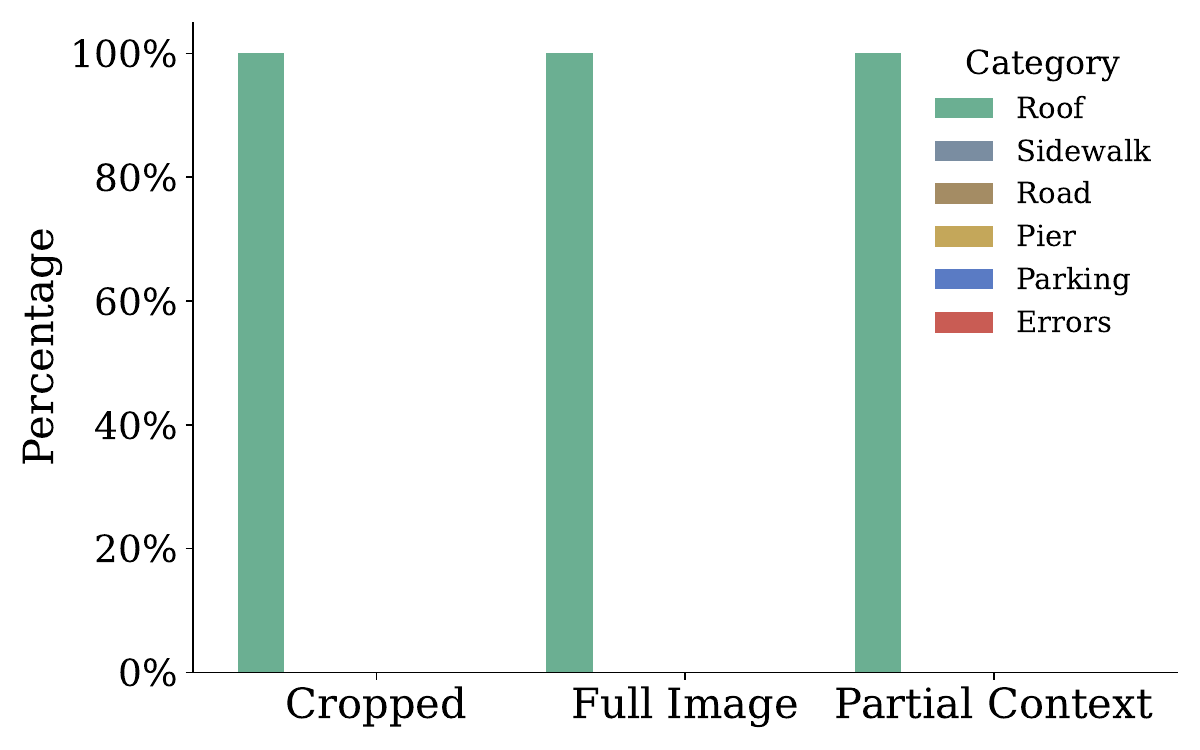}
         \caption{GPT-5 Mini}
    \end{subfigure}    
    \begin{subfigure}[b]{0.49\textwidth}
         \includegraphics[width=\textwidth]{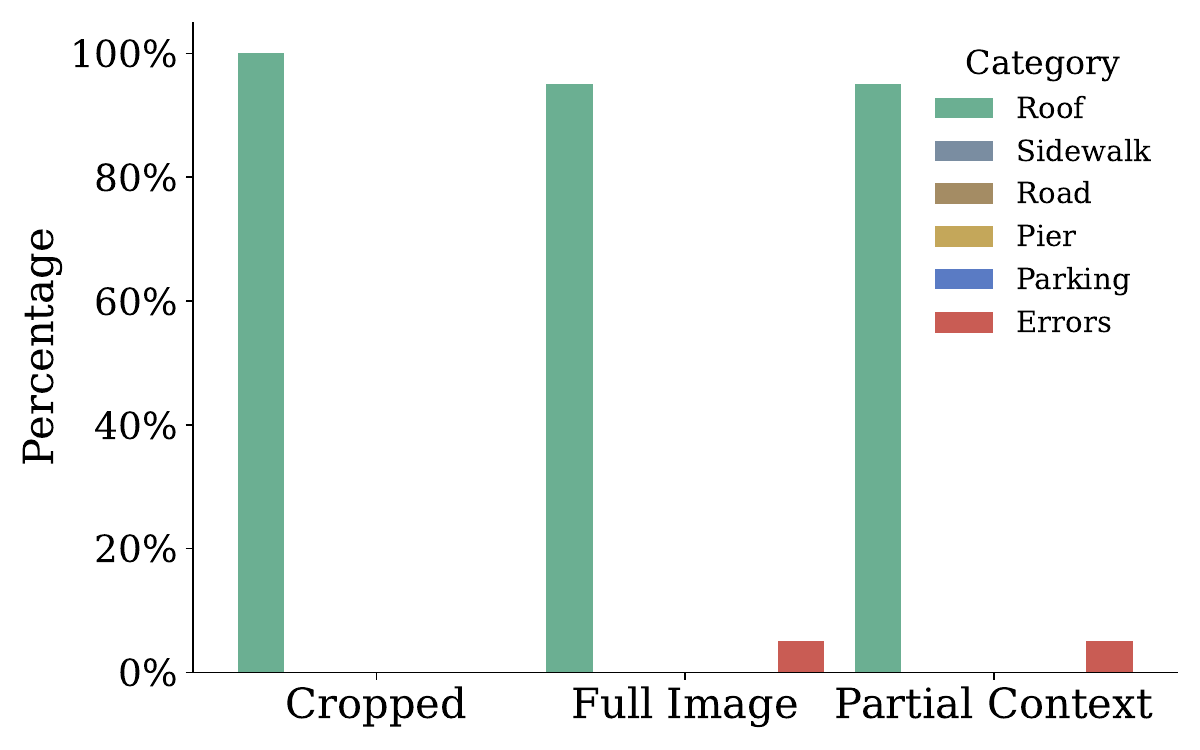}
         \caption{GPT-5 Nano}
    \end{subfigure}
    \begin{subfigure}[b]{0.49\textwidth}
         \includegraphics[width=\textwidth]{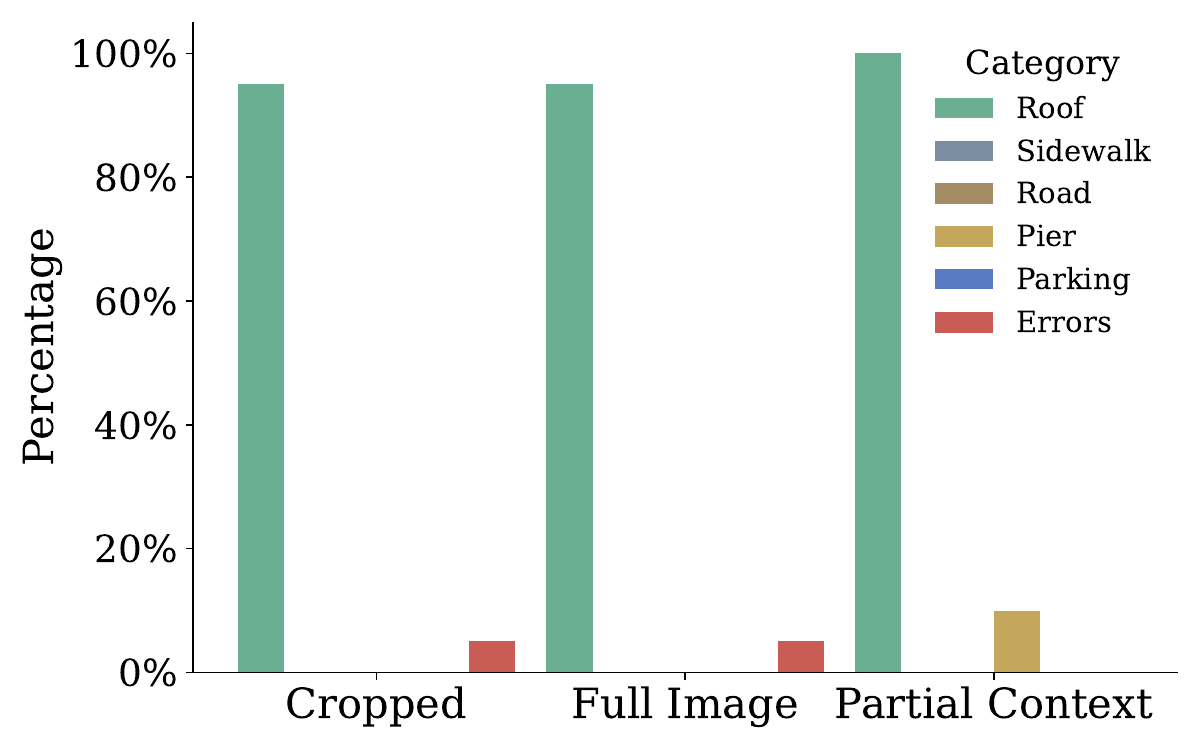}
         \caption{Gemini F 2.5}
    \end{subfigure}
    \begin{subfigure}[b]{0.65\textwidth}
         \includegraphics[width=\textwidth]{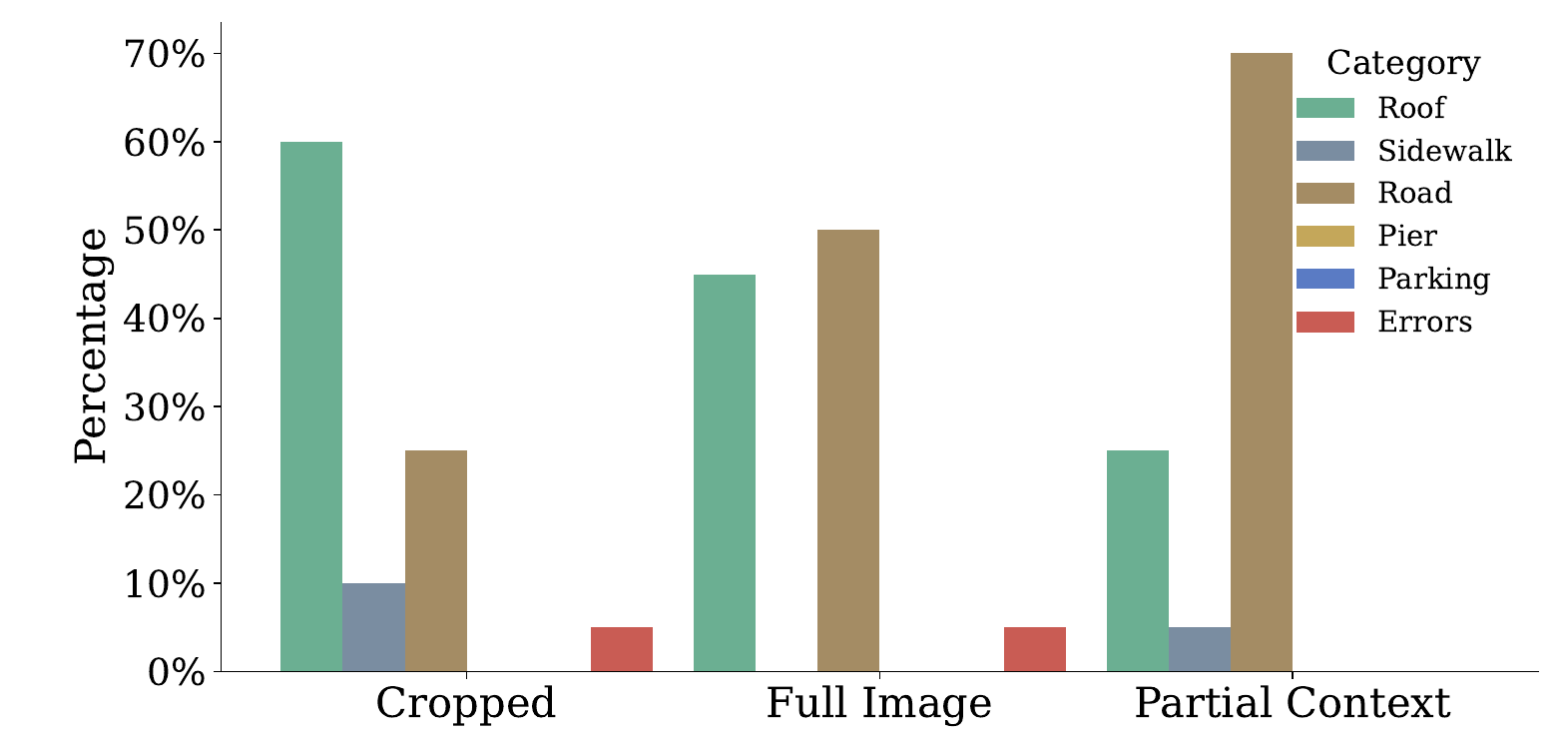}
         \caption{Mistral L 3}
    \end{subfigure}
    \caption{Scenario 2: Type of surface landed on by each model by information level} 
    \label{fig:surfaceType2}
\end{figure*}

\begin{figure*}[ht]
    \centering  
    
    \begin{subfigure}[b]{0.6\textwidth}
         \includegraphics[width=\textwidth]{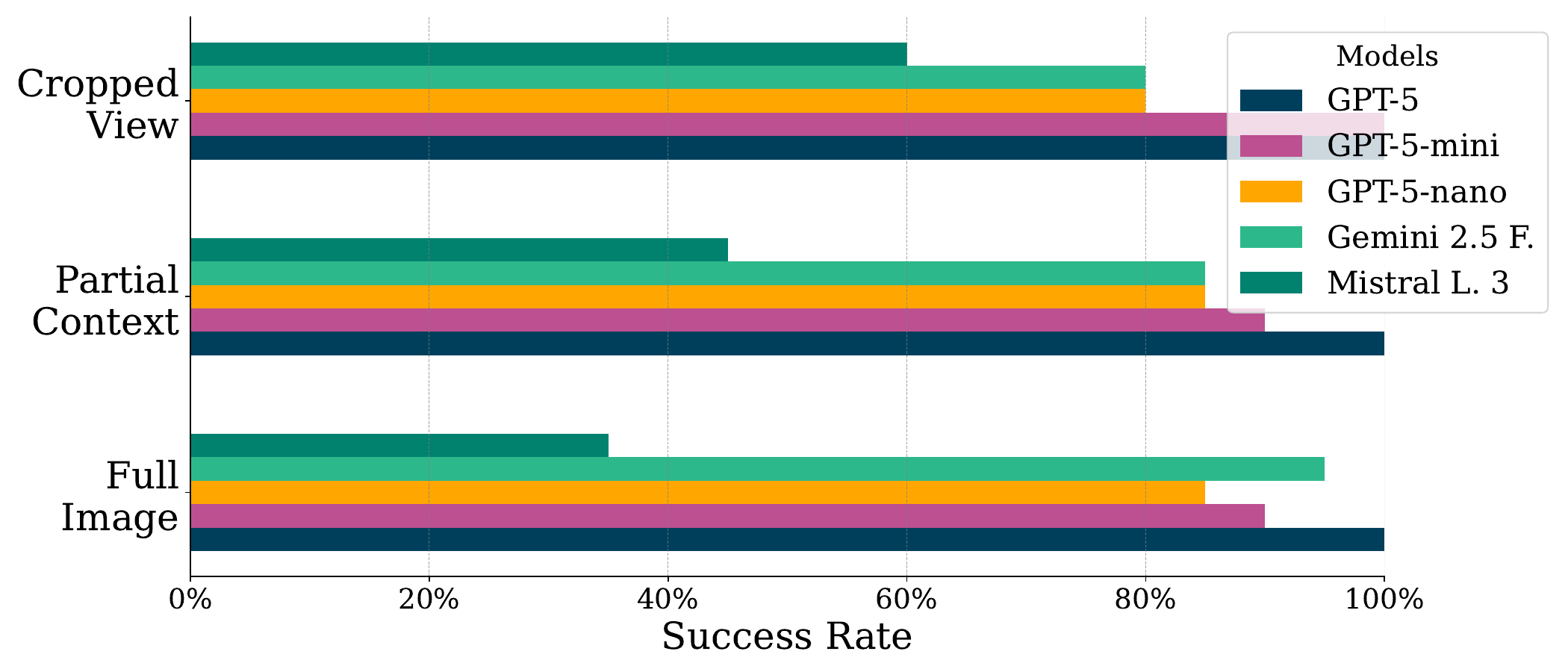}
         \caption{Scenario 1}
    \end{subfigure}
    \vspace{1em}
    \centering 
    \begin{subfigure}[b]{0.6\textwidth}
         \includegraphics[width=\textwidth]{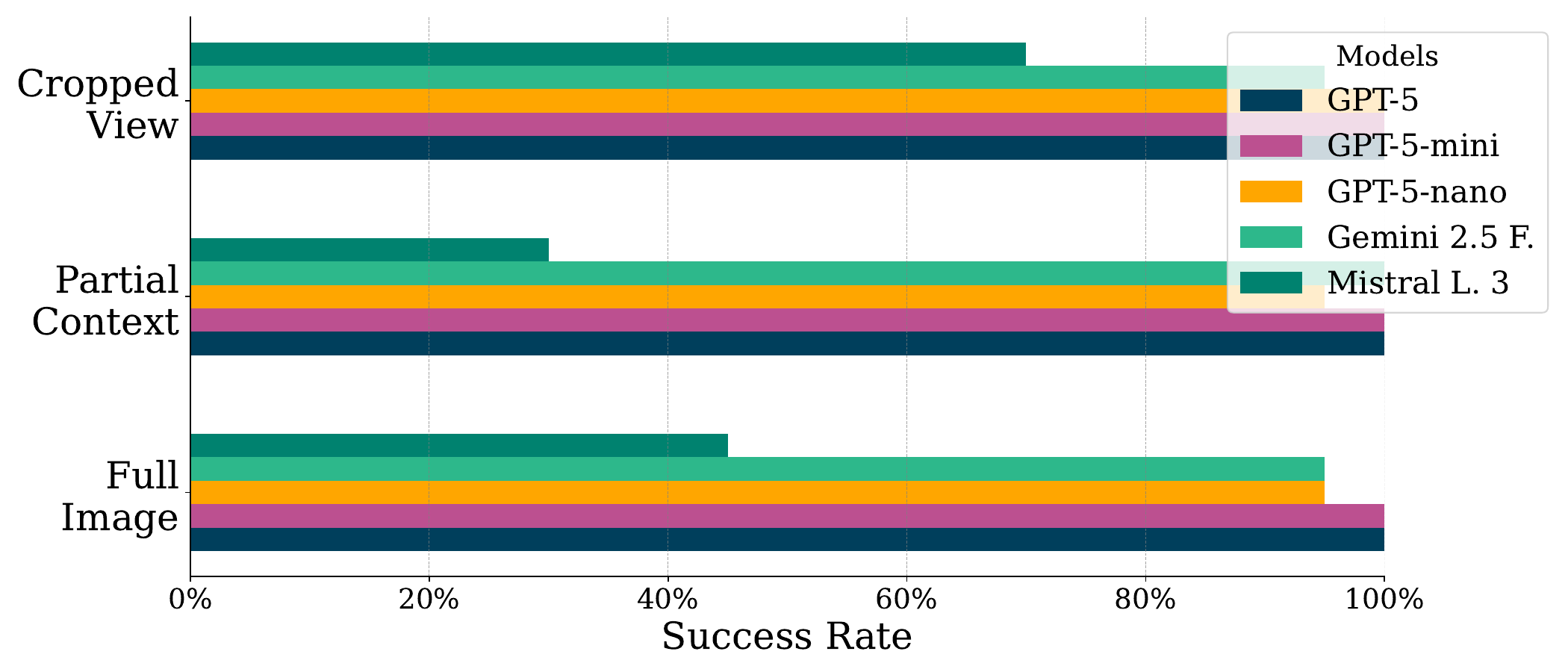}
         \caption{Scenario 2}
    \end{subfigure}    
    \caption{Pipeline performance per Scenario } 
    \label{fig:succPipeline}
\end{figure*}

With the individual modules validated, we next evaluated the full pipeline (running each module one after the other) in the two curated scenarios. Each run (20 iterations per scenario) started from the same initial position and tested different levels of context (cropped, partial 30\% padding, and full image). Unlike earlier tests, the full simulation exposed the pipeline to a broader variety of surfaces and dynamic conditions. We defined a landing as successful if the chosen surface was flat, free of people and objects, away from traffic, edges, and water, and therefore safe under real-world constraints. Due to hardware constraints we did not run the full pipeline on InternVL. 

Figures~\ref{fig:surfaceType1} and~\ref{fig:surfaceType2} categorize the types of surfaces selected by each model. GPT-5 consistently chose empty rooftops regardless of context, demonstrating robust reasoning. GPT-5-mini performed similarly with cropped inputs but showed more variability when additional context was provided, occasionally selecting less safe options such as roads or piers. GPT-5-nano displayed the widest variety of choices, performing best on rooftops with partial context but also producing errors, including hallucinations where it identified non-existent surfaces.

Similar to GPT-5 Mini, Gemini 2.5 F. performed best on cropped images, consistently landing on rooftops, with only a few exceptions. However, with partial context, performance diminishes, making the vehicle land on other types of surfaces and generating system errors. Mistral 3 L. had a diverse selection of surfaces across all levels of context, particularly favoring roads in Scenario 2. We can attribute this to its inability to distinguish between particular wall and roof textures. 

Applying the safety criteria, we computed overall success rates (Fig.~\ref{fig:succPipeline}). GPT-5 achieved perfect performance, while GPT-5-mini maintained success rates above 90\%, with occasional failures near piers. Nearby, Gemini 2.5 Flash performed 80\% or higher in all conditions across both scenarios, indicating an overall good performance with occasional landings in unsafe areas, such as roads, and system errors.Of the GPT models, GPT-5-nano performed worst, especially in Scenario~1, where shadows, textures, and rooftop walls caused frequent misclassifications. Scenario~2, with more uniform surfaces, was easier for Nano to handle, though it still lagged behind larger models. Mistral L. 3 performed the worst of all models, especially on Scenario 2, where most of the unsafe landings occurred around roads or highways that could include live traffic. These results highlight a clear trade-off: larger LVLMs are consistently reliable, while smaller ones require additional context and remain more prone to errors in complex urban settings.

When calculating resource consumption for these experiments, it show that our base pipeline consumes, on average, 715.629 MB of system memory and uses 0.31GB of GPU memory, with a peak allocation 0.41 GB. These results demonstrate that, despite using complex and intensive processes such as monocular depth-estimation, our pipeline remains lightweight enough while using cloud-based LVLMs to run on commercially available laptops without specialized hardware.

The controlled experiments with two scenarios served as an important validation step. By creating situations with an obvious safe landing option, we confirmed that the modules integrate smoothly, the LVLM can reliably recognize the intended choice, the pipeline can execute an end-to-end maneuver when the decision space is constrained and resource consumption is lightweight enough to run on an edge device. These results establish a baseline of correctness and give confidence in the system’s operation under simplified conditions, while also underscoring that such settings are only a first step toward the more complex realities drones encounter in practice.

\subsubsection{Pipeline performance under adverse weather conditions}
\begin{figure*}[ht]
    \centering  
        \hfill\begin{subfigure}[b]{0.49\textwidth}
         \includegraphics[width=\textwidth]{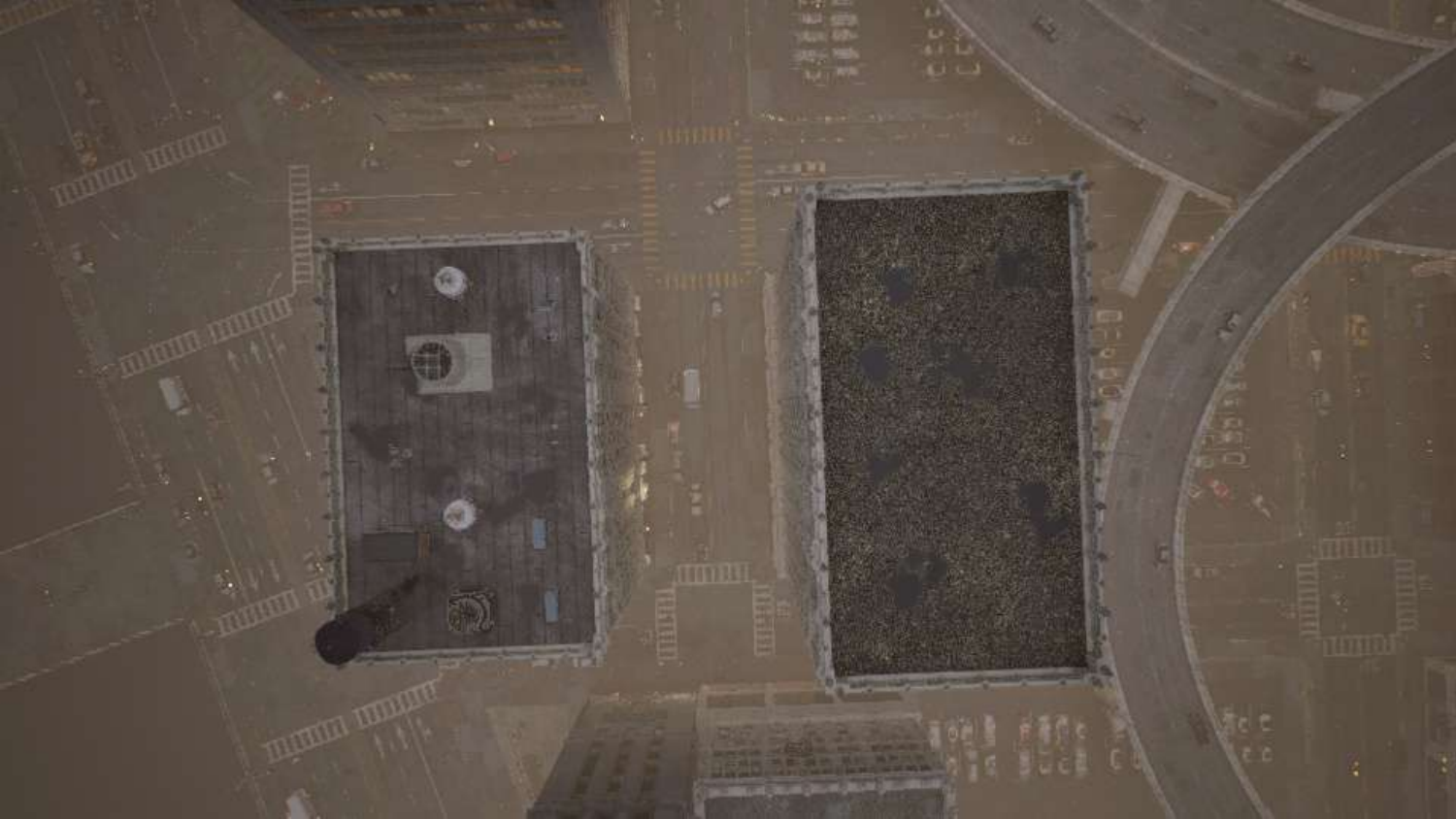}
         \caption{Scenario 1 under rain conditions}
    \end{subfigure}
    \hfill\begin{subfigure}[b]{0.49\textwidth}
         \includegraphics[width=\textwidth]{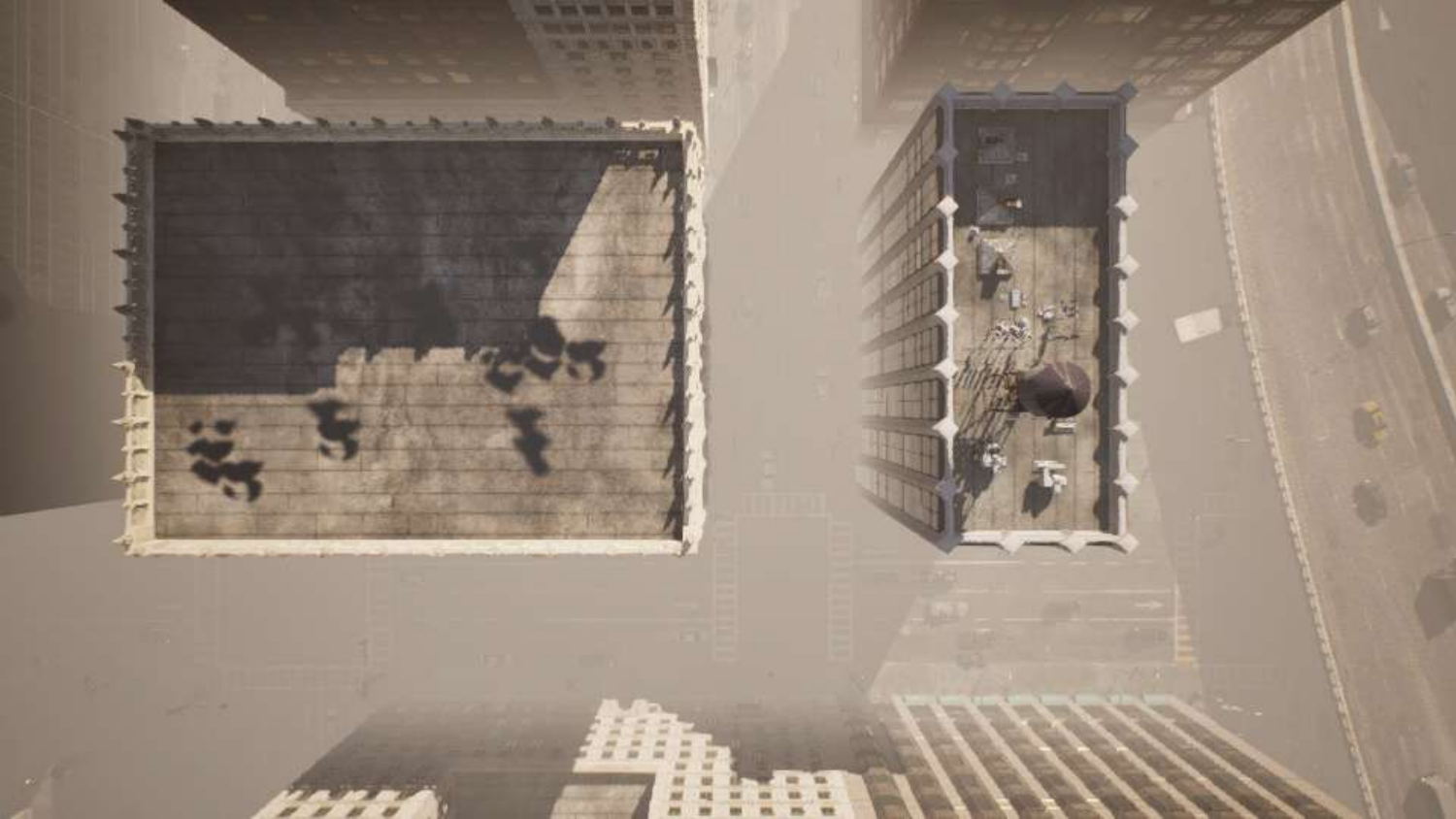}
         \caption{Scenario 2 under fog conditions }
    \end{subfigure}

    \caption{Adverse weather effects} 
    \label{fig:weather}
\end{figure*}
To evaluate the adaptability of our pipeline, we used Unreal Engine to extend our two previous scenarios and add adverse weather conditions, specifically fog and rain as shown in Fig. \ref{fig:weather}. We assess each scenario for 20 iterations using GPT-5 Gemini 2.5 F. and Mistral L. 3, following the same success criteria of the previous experiments. The results for Scenarios 1 and 2 are available in Tables \ref{tab:weather_perf1} and \ref{tab:weather_perf2} respectively. 

\begin{table}[h!]
\centering
\caption{Success Rate Under Adverse Weather Conditions (Scenario 1) }
\label{tab:weather_perf1}
\begin{tabular}{lccc}
\toprule
Condition & GPT-5& Gemini 2.5 F& \ Mistral L 3 \\
\midrule
Fog  & 100\% & 90\% & 30\% \\
Rain & 95\%  & 95\% & 50\% \\
\bottomrule
\end{tabular}
\end{table}

Across both scenarios, GPT-5 and Gemini 2.5 F. performed similarly successfully landing on safe surfaces 95\% of the time or more. In contrast, Mistral L. 3 showed inconsistent performance, improving during fog conditions but maintaining a poor record in the rain, with both scenarios having a 50\% or lower success rate. These results show that larger models are more robust to dynamic environmental changes, while smaller models can exhibit inconsistent behavior due to perception degradation, which could significantly alter mission success.

\begin{table}[h!]
\centering
\caption{Success Rate Under Adverse Weather Conditions (Scenario 2) }
\label{tab:weather_perf2}
\begin{tabular}{lccc}
\toprule
Condition & GPT-5& Gemini 2.5 F& \ Mistral L 3 \\
\midrule
Fog  & 95\% & 100\% & 90\% \\
Rain & 100\%  & 100\% & 35\% \\
\bottomrule
\end{tabular}
\end{table}

\section{Evaluation in a Realistic Urban Environment}
\begin{figure*}[ht]
    \centering  
\hspace{-5mm}
\includegraphics[width=1.05\textwidth]{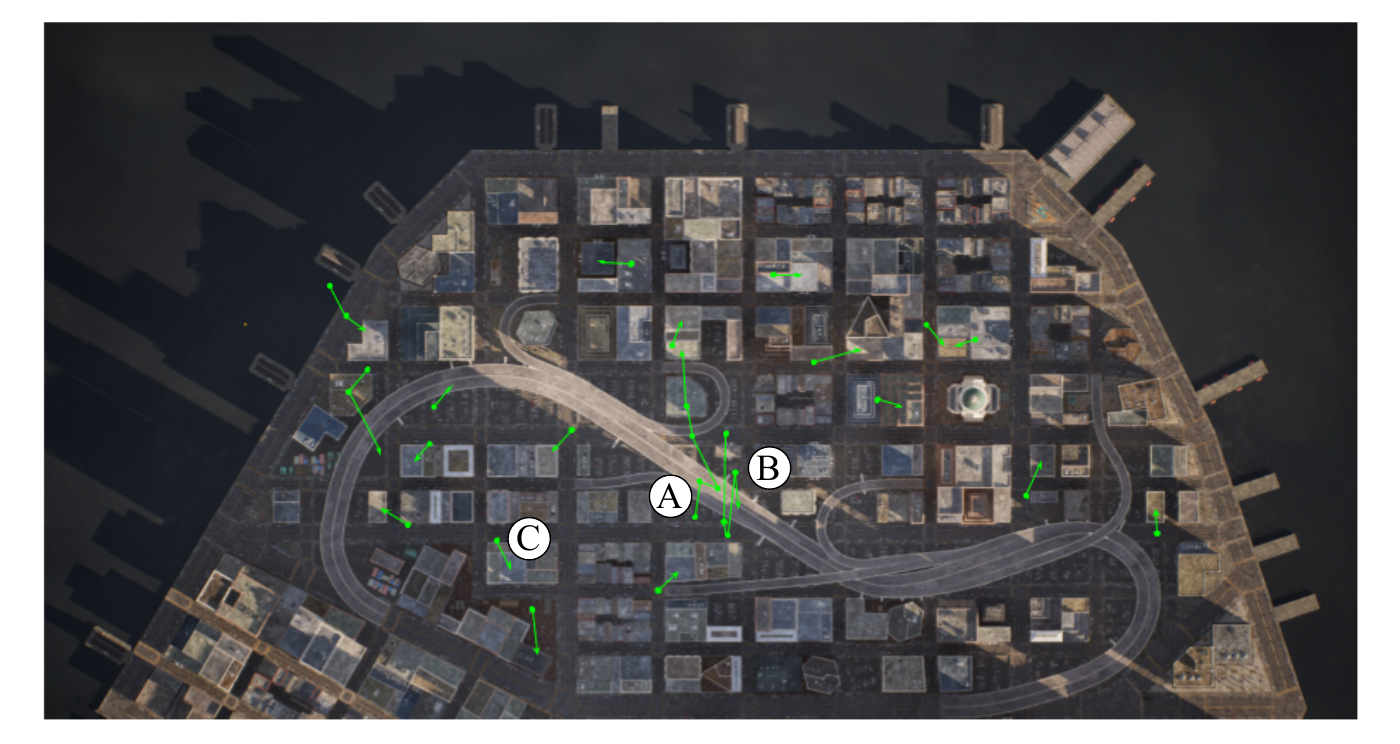}
         \caption{Bird's eye view of city with all drone trajectories}

    \label{fig:arrow_map}
\end{figure*}

The previous experiments validated our pipeline in curated scenarios where there was a clear and unambiguous safe landing choice. These settings were essential to confirm that the modules integrate correctly and that the LVLM can consistently select the intended option. Real cities, however, present a much harder problem: there is rarely a single “obvious” choice, and drones must instead reason about multiple possibilities that may all be partially safe or unsafe. Rooftops vary in slope and clutter, sidewalks and roads may appear flat but are risky due to traffic, and vehicles or pedestrians can suddenly appear in areas that initially seemed safe. The challenge in this section is therefore to evaluate how well our pipeline performs when it must navigate this ambiguity and make context-aware decisions without a predefined ground truth.

To keep the focus on system behavior rather than hyperparameter search, we use a set of parameters identified in the previous section with 100\% full-pipeline success rate (GPT-5, cropped view), and carry them forward unchanged into this open evaluation.

To structure this evaluation, we design a set of city-wide experiments that stress-test different aspects of the pipeline. To avoid biasing the results toward a handful of locations, we seeded drone launch sites using a two-dimensional Halton sequence~\cite{halton1960efficiency} (a quasi-random sequence with low discrepancy that has been applied in Monte Carlo simulations and falsification studies~\cite{dreossi2019compositional}) which distributes points evenly across the map while avoiding the clustering that can arise from purely random sampling. This ensures that the pipeline is exposed to a diverse range of urban contexts—dense blocks, open rooftops, highways, and mixed-use areas—rather than being confined to a few convenient spots.


We initialized the drone in 20 random locations selected by our Halton sampler. Figure~\ref{fig:arrow_map} shows the drone's initial positions, trajectories, and final landing positions, with dots indicating stop positions and arrows the direction of the movement and its final position.

With this setup, we ask three key questions: (i) how safe are the landings that result, including both successes and failure cases such as obstructed rooftops or highways; (ii) how quickly can the system converge to a landing decision, measured by the number of rounds before descent;  and (iii) what reasoning patterns does the LVLM exhibit, and how do they vary across different contexts? Together, these experiments provide a comprehensive picture of the pipeline’s behavior in a dynamic city environment, highlighting both its strengths and its current limitations.


\subsection{Safety Analysis}
We study the 20 scenarios and qualitatively analyze whether the final landing area can be considered safe. Our safety assessment is a combination of areas without major obstacles, people, or other hazards such as vehicles. 

\begin{figure*}[ht]
    \centering  
    
    \hfill\begin{subfigure}[b]{0.49\textwidth}
         \includegraphics[width=\textwidth]{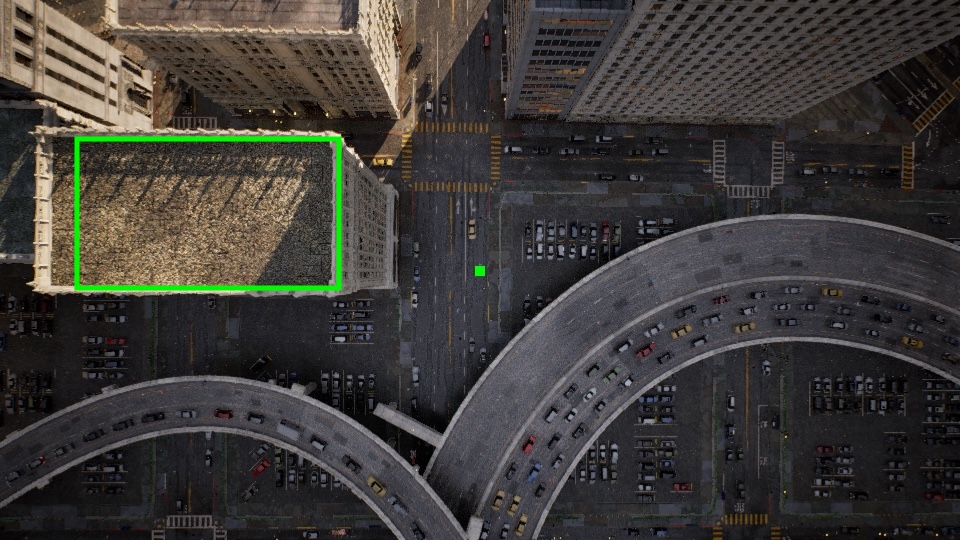}
    \end{subfigure}
    \hfill
    \begin{subfigure}[b]{0.49\textwidth}
         \includegraphics[width=\textwidth]{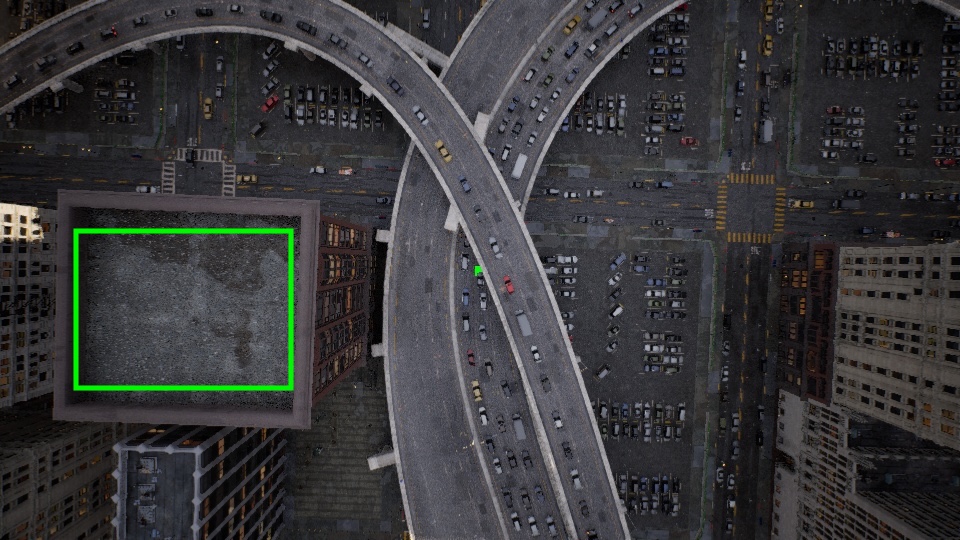}
    \end{subfigure}    
    \caption{Examples of successful landing scenarios.} 
    \label{fig:good_zone}
\end{figure*}

During 20 iterations, the drone managed to land on a safe surface in 75\% of the cases. All of these cases occurred on open rooftops, as illustrated in Fig.~\ref{fig:good_zone}. 

We observed five problematic outcomes:

\paragraph{Obstructed rooftops (2 cases).} The LVLM cleared landings on rooftops with dense HVAC structures. Although these surfaces were geometrically flat, clutter increased collision risk and jeopardized relaunch, suggesting that our prompts should penalize rooftop clutter more aggressively.

\paragraph{Highways (2 cases).} The drone landed on a highway twice. In one instance, no traffic was visible at decision time, but the roadway remained unsafe due to potential traffic reappearance; in the other, a vehicle was already present and alternatives were heavily obstructed. This indicates we should encode stronger relative risk between “flat but risky” (roads) and “flat and benign” (rooftops, parking rooftops).

\paragraph{Timeout (1 case).} The pipeline exceeded the maximum rounds (10 in our case) and forced a landing after repeatedly rejecting otherwise safe sites; a misclassification near the end (the LVLM confused pipes for wildlife, and decided that in order to protect the birds, it was not going to land on an otherwise clear rooftop).

\subsection{Number of Rounds Needed to Complete a Landing}

A \emph{round} is one full pipeline loop (Fig.~\ref{fig:pipeline}): surface detection $\rightarrow$ LVLM ranking $\rightarrow$ movement $\rightarrow$ LVLM confirmation. Fewer rounds imply faster and more decisive behavior in a dynamic scene.

\begin{figure*}[ht]
    \centering  
    \includegraphics[width=0.5\textwidth]{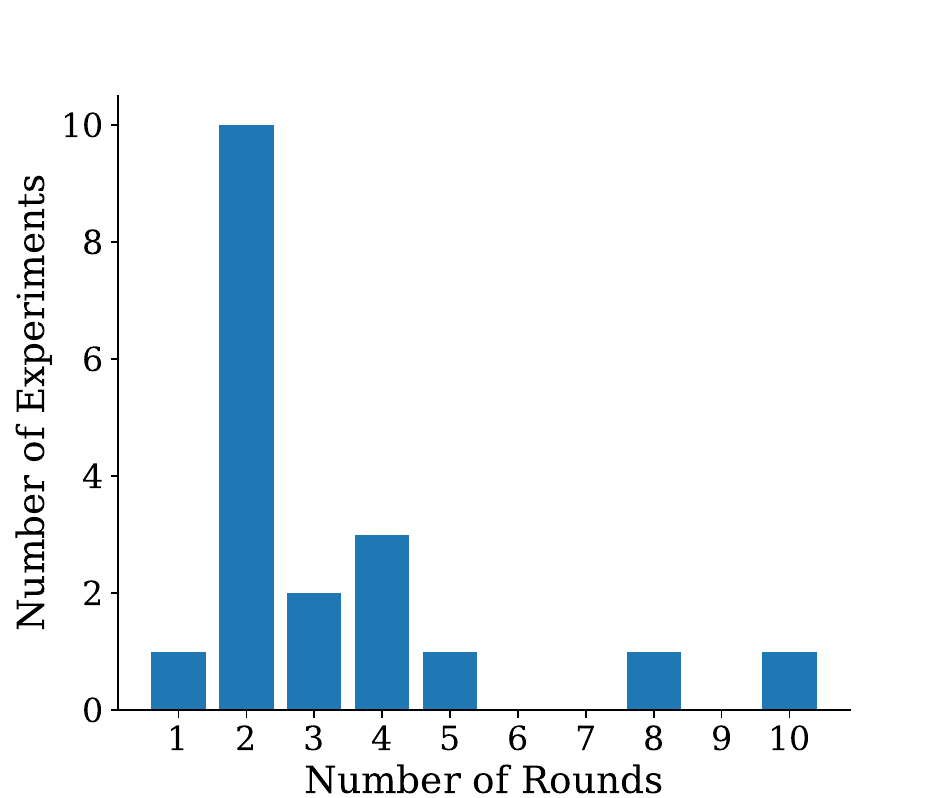}
         \caption{Number of rounds needed for landing.}
    \label{fig:rounds}
\end{figure*}


As shown in Figure~\ref{fig:rounds},  more than 50\% of the experiments concluded within two rounds, typically on nearby rooftops (e.g., marker \textit{C} in Figure~\ref{fig:arrow_map}).

\begin{figure*}[ht]
    \centering  
    \includegraphics[width=\textwidth]{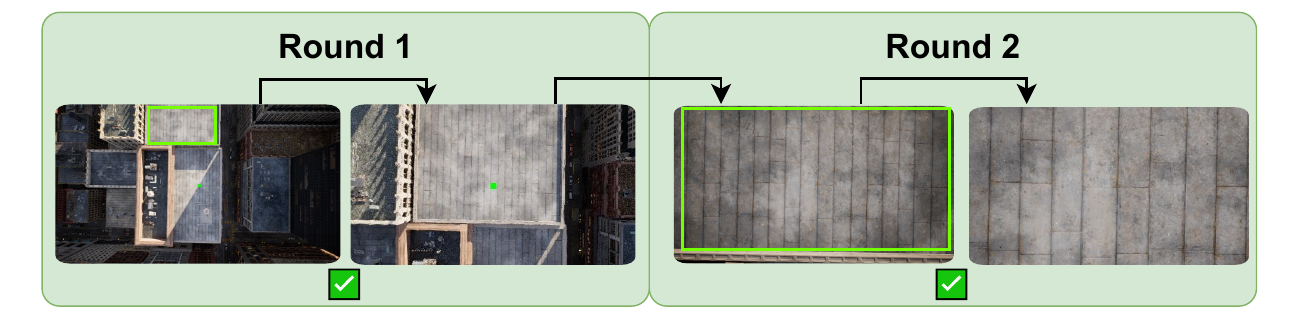}
         \caption{Most Common Example.}
    \label{fig:basicRounds}
\end{figure*}

Most of our safe landings take on average 2.2 rounds, indicating that our process can act quickly and correctly in dynamic environments. The most common example of our pipeline is illustrated in Figure~\ref{fig:basicRounds} where the drone drone identifies and navigates to a safe landing surface in the first run while confirming and arriving at the surface in the second round.

\subsection{Edge Cases}

\noindent \textbf{Edge Case 1:} We now explore some of the edge cases that required multiple rounds of the pipeline. 

\begin{figure*}[ht]
    \centering  
    
         \includegraphics[width=\textwidth]{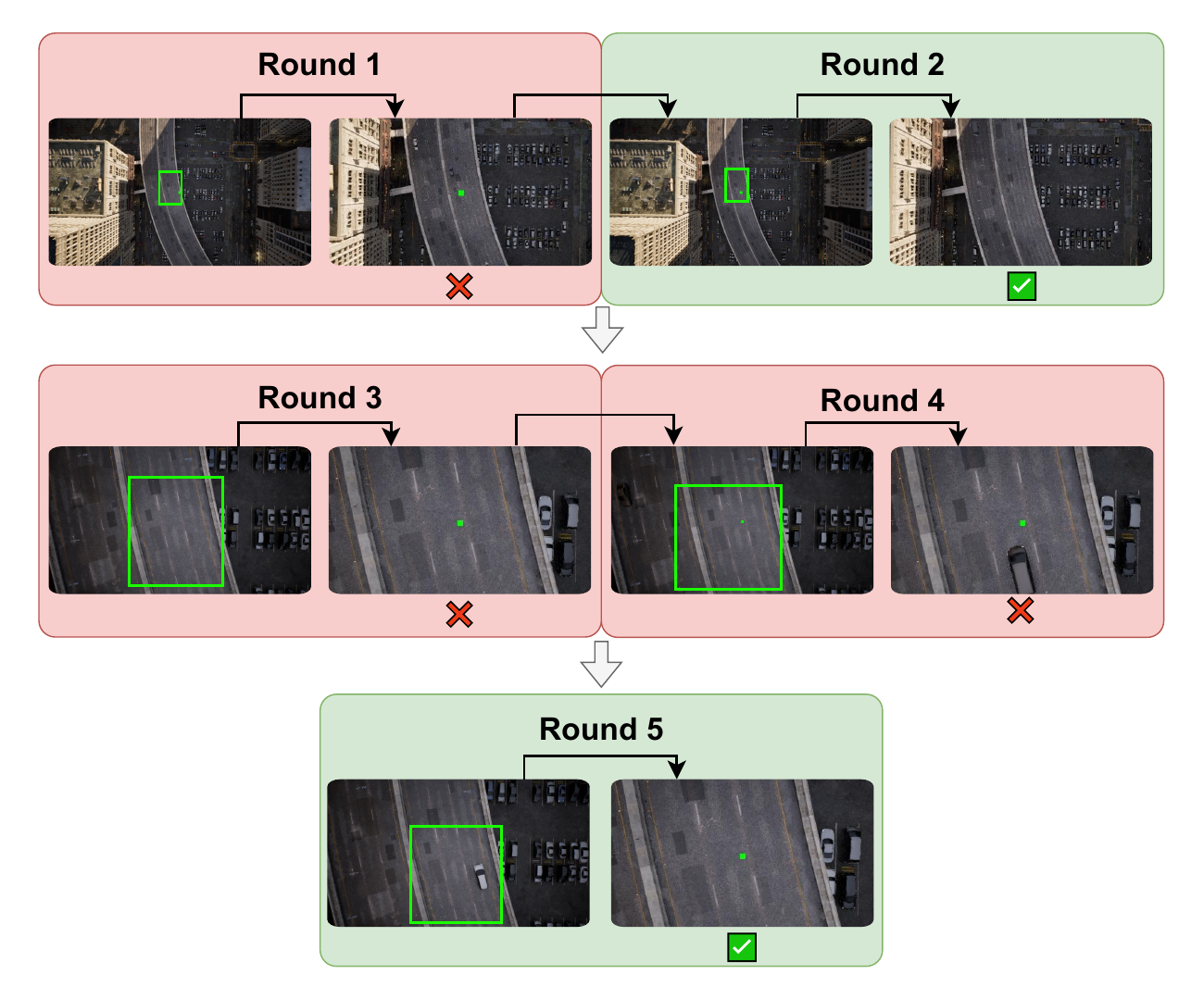}
         \caption{Hesitation for Landing on a Highway.}

    \label{fig:dynamic_comic}
\end{figure*}

Marker \textit{B} in Fig.~\ref{fig:arrow_map} highlights a launch over a sparse area with few suitable zones in view. Figure~\ref{fig:dynamic_comic} shows all five rounds. 

In this scenario, the drone experiences multiple failed attempts due to two primary factors: (1) dynamic changes in the landing environment between detection and approach phases, and (2) inconsistent safety assessments by the LVLM. This scenario demonstrates the challenges inherent in real-time autonomous decision-making systems operating in dynamic urban environments. Below we pair the LVLM’s decisions with the key quotes from the LVLM.


\textbf{Round 1:} The LVLM initially identified a highway segment as the optimal landing zone, reasoning that the absence of traffic outweighed the explicit instruction to avoid highways, particularly given that alternative zones contained excessive obstacles. 

\begin{quote}
    \texttt{Wide elevated roadway appears clear of vehicles and pedestrians in the landing zone. Surface is flat and unobstructed, offering ample space. Risk remains from potential oncoming traffic and guardrails at the edges, but within the frame it meets the constraints best.}
\end{quote}
However, during the transit time to the proposed landing site, a vehicle entered the highway. Upon reaching the confirmation stage, where the drone reassesses landing zone safety from its new position directly above the target, the LVLM correctly identified the presence of the vehicle and aborted the landing attempt.
\begin{quote}
    \texttt{ \raggedright Not suitable for final approach. The marked spot is in an active traffic lane on a roadway. There are vehicles nearby, making the area not clear of obstructions and posing collision risk. Although the road appears generally flat, it is not clear of traffic/obstacles and is near roadway edges/structures, which further increases risk.}
\end{quote}

\textbf{Round 2:} As the LVLM does not confirm the landing, we run the pipeline again. The system again selected the same highway segment, which had cleared of traffic between rounds. During the confirmation stage, the highway remained clear, prompting the drone to initiate a controlled descent. However, as the drone had not yet reached sufficient proximity for final landing, another decision round was triggered.

\textbf{Round 3:} The LVLM identifies the highway once more as the safest option. Upon reaching the confirmation stage, the physical environment remained unchanged, with a clear highway.
\begin{quote}
    \texttt{ \raggedright Best option. It shows a wide, flat asphalt roadway with clear lane markings. The central lanes appear unobstructed and free of people and vehicles; parked cars are confined to the right-side parking bay behind a curb. Surface looks level and continuous, with only minor patches/markings that don’t pose hazards. Risk remains from potential unseen traffic, but among the options this is the widest, clearest flat area.}
\end{quote}

\textbf{Round 4:} Similar to Round 1, the drone initially assessed the highway as safe, but vehicular traffic appeared during the transit period. The confirmation stage correctly identified this new obstacle, leading to another cycle repetition.

\textbf{Round 5:} Despite the presence of a vehicle on the highway, the LVLM determined this location remained the safest available option, as the alternative parking lot was deemed too cluttered with obstacles for safe landing. Fortuitously, the highway cleared during the drone's approach, enabling successful confirmation and completion of the landing sequence.

\begin{quote}
    \texttt{ \raggedright The highlighted area is on a wide, flat roadway. The surface appears level with no visible debris or protrusions near the target. There are no people in the landing zone. Parked cars and the curb/guardrail are off to the right but not within the immediate touchdown area, leaving adequate clearance if you stay centered between lane markings. Visually, this meets the criteria for a clear, flat surface, though remain vigilant for any approaching vehicles during descent.
}
\end{quote}

\noindent\textbf{Takeaways.} Safe choices can become unsafe during motion. Although the highway might be empty at some point in time, it becomes unsafe at others. As the LVLM correctly identified in the first round, ``Risk remains from potential oncoming traffic'', so this is something we should emphasize in future prompts; if the future risk is high, we should ignore that surface and search for better alternatives in other places, and not keep hesitating on whether or not to land on a highway. 

\begin{figure*}[ht]
    \centering  
    \includegraphics[width=\textwidth]{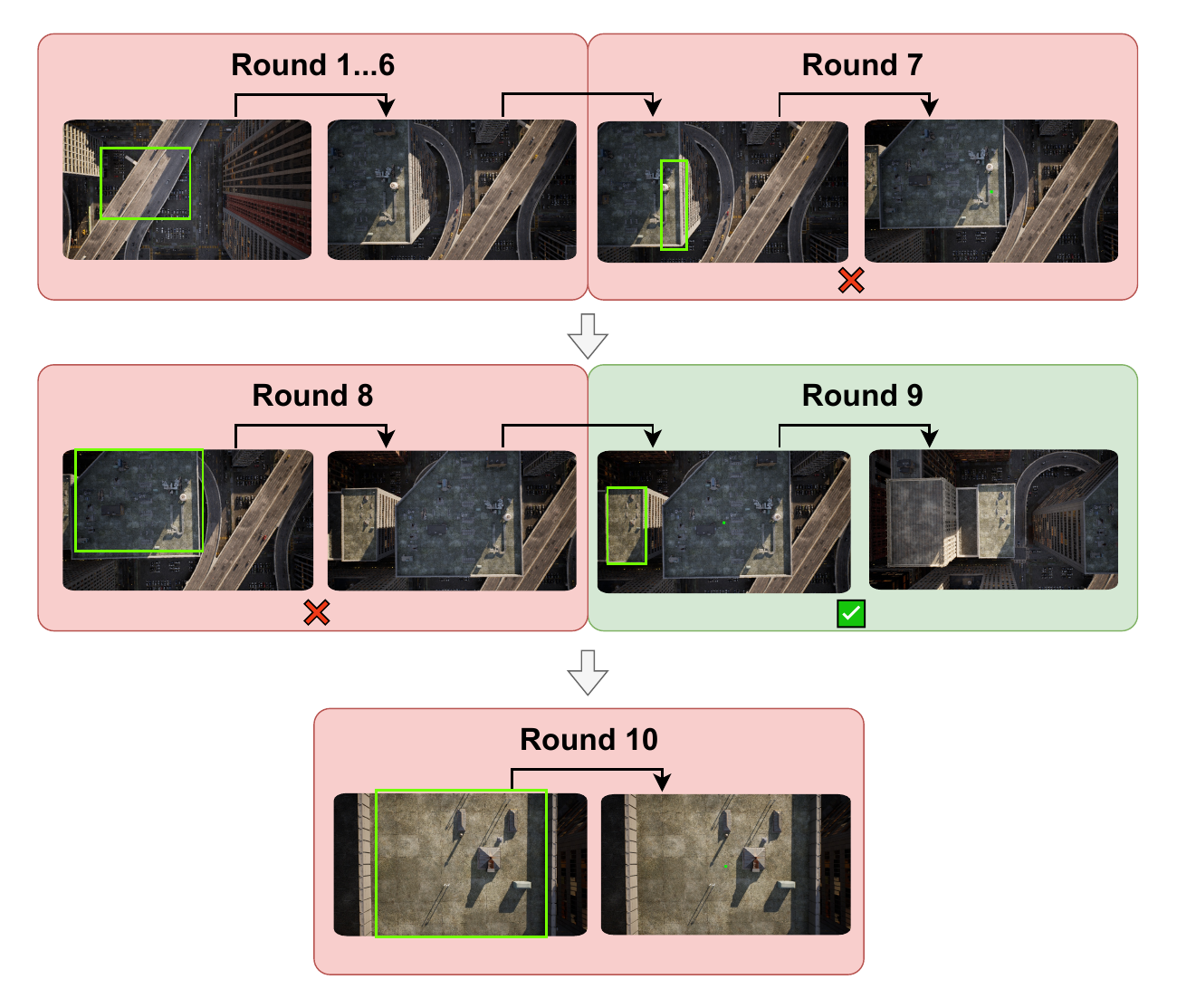}
         \caption{Failure due to timeout}

    \label{fig:failedRounds}
\end{figure*}
\noindent \textbf{Edge Case 2:} We now analyze the case where we reached 10 rounds without landing. Marker \textit{A} (Fig.~\ref{fig:arrow_map}) shows this run.  We illustrate the different rounds in Fig.~\ref{fig:failedRounds}.

We can attribute the failure of this experiment to (1) the limited availability of landing spots in the first rounds given to the LVLM by our surface detector and (2) the misclassification of an object as a bird in the final round.

\textbf{Round 1-6:} In these rounds, the LVLM detected different parts of the same highway moving slightly between lanes but failing to confirm due to the presence of vehicles moving beneath it and moving at the end of Round 6 to an off-ramp from the highway, managing to change the camera view significantly. 
\begin{quote}
        \texttt{ \raggedright Not suitable for final approach. The scene shows an active roadway and a crowded parking lot. There are many vehicles (moving on the street and parked in rows), leaving no sufficiently large, obstacle‑free patch. Any gaps between cars are too tight for safe rotor clearance. The elevated road surface is also a traffic route with edge hazards and uncertain vehicle presence. While surfaces appear flat, they are not clear of obstructions and present collision risks.}
\end{quote}

\textbf{Round 7}: In this round, the LVLM selected a fraction of the adjacent rooftop as a suitable surface, finally leaving the highway area and positioning itself on top of a rooftop. However, the LVLM did not confirm the final approach due to the edge and obstacles on the roof.

\begin{quote}
        \texttt{ \raggedright The rooftop appears flat and there are no people visible, but the selected spot is not safe. It lies very close to multiple obstructions (HVAC pipes/ducting and a vent) and is also near the roof edge/parapet. These increase collision risk and potential downdraft/edge turbulence during landing.}
\end{quote}

\textbf{Round 8}: In this round, the LVLM selected the complete rooftop as a possible landing spot, proceeding to its center, recognizing its possible obstacles and failing to confirm because of them.

\begin{quote}
        \texttt{ \raggedright The roof appears flat and there are no people or vehicles. While there are many HVAC units, vents, and pipes, there are multiple open gravel areas large enough for a quadcopter to touch down, especially in the lower-left and central-left zones away from equipment and roof edges.}
\end{quote}

\textbf{Round 9}: With a better position, the system managed to detect an alternative landing surface at another roof further to the left, and unlike previous iterations, the LVLM cleared the drone for a descending operation.

\begin{quote}
        \texttt{ \raggedright Proceed with final approach to a clear section of this roof, avoiding the central vent cluster and maintaining margin from edges.}
\end{quote}

\textbf{Round 10}: In its final round, the drone identified the roof beneath it as the best option. However, it failed to confirm the landing, justifying its rationale on the presence of nonexistent wildlife on the roof. 
\begin{quote}
        \texttt{ \raggedright However, there are two birds on the roof very close to the proposed touchdown point. Moving wildlife is an immediate obstruction risk during landing (possible collision or sudden movement into the rotors).}
\end{quote}

\noindent\textbf{Takeaways.} (i) When nearby safe surfaces are scarce, the agent can loop locally before exploring further; (ii) incorrect object identification can veto otherwise safe landings—motivating additional confirmations or sensor cross–checks.

Overall, the big takeaway is that when there are no viable candidates in the first round, the drone should explore different areas and not settle for what it was originally perceived. Of course this decision might depend on the criticality of the alert that requires the drone to land. If it has to land immediately perhaps a risky situation is more suitable than attempting to move somewhere else.

\subsection{Reasoning Analysis}

Finally, we study the reasons the LVLM provides to select a safe place to land, and understand the relevant elements in the decision.

We measure the semantic similarity of the LVLM final ranking responses, and then we embedded each response with OpenAI's \texttt{text-embedding-3-small}. This model allows us to convert the natural text output of the LVLM to a numerical representation (a vector).  We then perform clustering of these vectors to identify patterns of decisions.

\begin{figure*}[ht]
    \centering  
    
         \includegraphics[width=\textwidth]{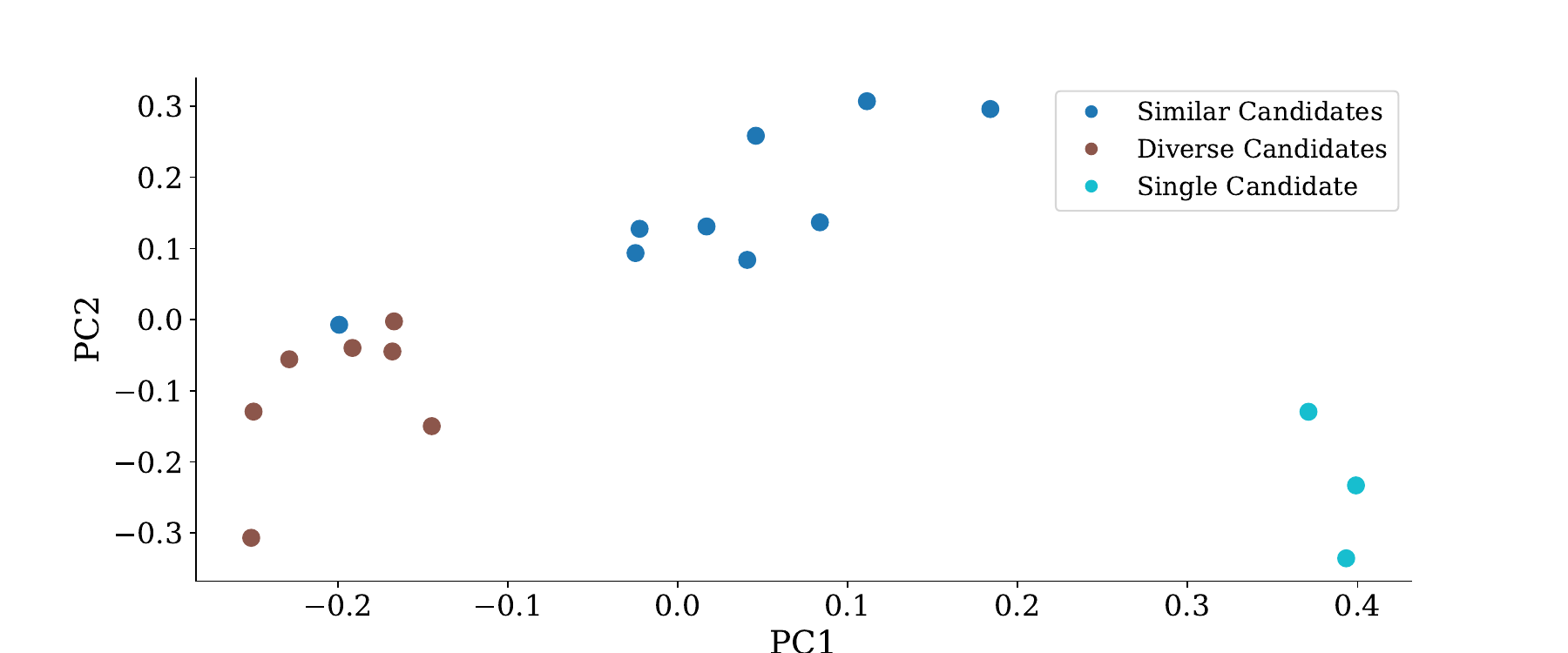}
         \caption{Final Decision Clusters }

    \label{fig:clusters}
\end{figure*}

A visualization of these vectors using Principal Component Analysis (PCA)~\cite{Jolliffe2016PrincipalCA}  is shown in Figure \ref{fig:clusters}. This figure suggests three clusters. 

By inspecting the output of the LVLM in each cluster, we suggest the following three categories of LVLM reasoning: \textbf{Similar Candidates:} The largest cluster includes cases where the LVLM described multiple candidate surfaces as broadly alike, often using more generic descriptions of the terrain without explicitly identifying the specific type of surface. \textbf{Diverse Candidates:} This cluster covers situations in which the LVLM has to make a selection over a broader variety of surfaces (e.g., roads, sidewalks, roofs), making these scenarios more complex than the previous cluster. \textbf{Single Candidate} This cluster contains instances that only deal with a single detected surface, where reasoning is more straightforward and focused on acknowledging minor risks while ultimately following the prompt to select the sole option available.

These clusters illustrate that the LVLM adapts its reasoning style to the complexity of the decision space: when candidates look similar, it falls back on generic comparisons; when options differ substantially, it weighs trade-offs more carefully; and when only one option is present, it performs a simple risk check before proceeding. This variability suggests that the model is not applying a fixed heuristic but is instead modulating its rationale based on context, an encouraging sign for embodied AI operating in open environments. At the same time, the analysis highlights opportunities for improvement: more explicit guidance could help the model be less generic when faced with similar candidates, and better calibration is needed when evaluating diverse options with unequal risk profiles. In future work, we will try to focus on identifying the safety arguments the LVLM makes.





\subsection{Lessons Learned}


Our experiments taught us several valuable lessons about creating effective LVLM prompts and input strategies. For one, we discovered that when we explicitly told the model to both rank candidate surfaces and explain its reasoning, it became more reliable. Prompts that focused on safety constraints—like requiring surfaces to be clear of obstacles, people, and non-flat areas—led to more structured outputs and reduced the chances of missing hazards. 

Second, we found that adding more context to the input did not always boost performance. Although we initially thought that including the entire scene would enhance reasoning, the most accurate results came from using only the cropped candidate surfaces. A key takeaway is that carefully curating the list of potential safe landing spaces can prevent the LVLM from getting overloaded with options and information. 

Lastly, we need to account for cases where no suitable surfaces are found. Although our implementation reran the pipeline in these cases, alternative designs could use a set of predefined locations to evaluate or heuristics to enhance robustness and address uncertainty. We also observed that compact models like GPT-5-nano show varying sensitivities to context across different model variants, underscoring the broader point that efficiency-focused LVLMs may sacrifice accuracy for speed or easier deployment.

\section{Conclusions and Future Work}

We introduced a hybrid recovery pipeline that leverages LVLMs for high-level reasoning while grounding decisions in conventional perception and control modules. This modular design proved effective: the system can identify and execute safe landings in realistic city environments, demonstrating the promise of embodied AI for adaptive recovery. At the same time, our experiments revealed a key limitation—reliable performance currently requires large models, which are difficult to host directly onboard resource-constrained drones.

This tension between capability and deployability motivates future work on hierarchical inference architectures. A promising approach is to combine lightweight, onboard geometric checks with mid-sized distilled models at the edge and large LVLMs in the cloud. Such a design could balance efficiency, reliability, and resilience to connectivity constraints. Another direction is to explicitly model and quantify uncertainty in LVLM decisions~\cite{bhatt2024know}. Being able to detect when the model is unsure would enable active sensing or fallback heuristics, providing additional safeguards in dynamic environments.

Finally, our study highlights the need to evaluate robustness against adversarial and deceptive inputs~\cite{mbv2025cvpr}. LVLM-based recovery systems remain vulnerable to visual perturbations such as phantom or patch attacks, and integrating adversarial resilience into the design will be essential for safety-critical deployments. Together, these directions chart a path toward recovery pipelines that are not only effective in simulation but also trustworthy and practical in real-world operations.

\section*{Acknowledgments}

This material is based on work supported in part by the Air Force Office of Scientific Research (AFOSR) under award number FA9550-24-1-0015, and by the National Center for Transportation Cybersecurity and Resiliency (TraCR) (a U.S. Department of Transportation National University Transportation Center) under grant numbers 69A3552344812 and 69A3552348317. Any opinions, findings, conclusions, and recommendations expressed in this material are those of the authors and do not necessarily reflect the views of TraCR, and the U.S. Government assumes no liability for the contents or use thereof.

\clearpage

\appendix
\section*{Prompts}

\textbf{System Prompt:}
\begin{quote}
    \ttfamily
    \#\#Safe Drone Landing Assistant

   You are a safety assistant for a quadcopter that is flying over a city and needs to perform an emergency landing on a surface. 
   The quadcopter will provide you with photos of several possible surfaces taken by its camera bellow, and ask questions of the suitability of the surfaces to perform the safest emergency landing possible.
   To classify as safe a potential landing site MUST take into consideration the following factors:

\#\# Constraints           

    - **MUST** be clear of any obstructions such as air ducts, cars, rubble etc.
    - **MUST** be clear of people 
    - **MUST** be a flat surface

\#\# Clarifications 
           
    - Ignore non-critical visual features such as shadows, surface textures, tile patterns, or color variations unless they **CLEARLY** indicate an actual obstruction.
    - Do not assume a surface is unsafe just because it might be something like a shingled roof unless there is strong visual evidence of danger (e.g., visible slope, fragile material, obvious gaps).
    - If no surface is perfectly safe, select the one with the lowest risk.
    - You **MUST** always return a clear selection (never refuse).
\end{quote}

\noindent\textbf{Ranking Prompt:}

\begin{quote}
    \ttfamily
    You are receiving up to **5 images** from the drone:

    - **Images 0 - 4** (maximum 5 images): Candidate surfaces for landing. 

    \#\#\# Task
    1. Analyze each candidate surface (Images 0–4).  
    2. Rank the candidate surfaces from **most suitable** to **least suitable** for landing.  
    3. Provide detailed reasoning for each candidate, explaining why it is more or less suitable. 
    4. Output the ranking as a list of indices (corresponding to the candidate images only).  

    \#\#\# Output Format
    1. **Detailed Explanation** (paragraphs describing reasoning for each candidate).  
    2. **Final Ranking**: output the indices corresponding to each photo, representing the ranking from the most suitable place to land to the least one. The index starts at 0. 

\end{quote}

\noindent\textbf{Confirmation Prompt:}

\begin{quote}
    \ttfamily
    You are receiving a **close-up image** of a potential landing surface that was previously selected by you.  
    Your task is to confirm whether this surface is safe for a **final landing approach**.

    \#\#\# Task
    1. Analyze the surface in the close-up image.  
    2. Decide if the surface is **safe to land** or **unsafe**.  
    3. Provide a detailed explanation of your reasoning.  
    4. Output your decision as an index array with a single value:  
        - 1 in the indices array to confirm or 0 to cancel the final approach.

    \#\#\# Notes
    - Ignore **non-critical details** such as shadows, lighting variations, or surface patterns unless they clearly indicate a real hazard.  
    - Confirm landing if the surface appears **flat, unobstructed, and free of people**.  
    - Deny landing only if there is clear evidence of danger (obstructions, slope, rubble, fragile/unsafe material).  
    - Always return a decision (never leave it blank).  

    \#\#\# Output Format
    1. **Explanation** (paragraph justifying your decision).  
    2. **Final Decision** in the form: 1 in the indices array to confirm or 0 to cancel the final approach.
\end{quote}

\bibliographystyle{ws-book-har}    
\bibliography{ws-book-sample}      

@article{turkcan2024boundless,
  title={Boundless: Generating Photorealistic Synthetic Data for Object Detection in Urban Streetscapes},
  author={Turkcan, Mehmet Kerem and Li, Yuyang and Zang, Chengbo and Ghaderi, Javad and Zussman, Gil and Kostic, Zoran},
  journal={arXiv preprint arXiv:2409.03022},
  year={2024}
}

@inproceedings{insaurralde2022ontology,
  title={Ontology-Based Situation Awareness for Air and Space Traffic Management},
  author={Insaurralde, Carlos C and Blasch, Erik and Sabatini, Roberto},
  booktitle={2022 IEEE/AIAA 41st Digital Avionics Systems Conference (DASC)},
  pages={1--8},
  year={2022},
  organization={IEEE}
}

@inproceedings{zhang2020rtss,
  author    = {Lin Zhang and Xin Chen and Fanxin Kong and Alvaro A. Cardenas},
  title     = {Real-Time Attack-Recovery for Cyber-Physical Systems Using Linear Approximations},
  booktitle = {Proceedings of the 41st IEEE Real-Time Systems Symposium (RTSS)},
  year      = {2020},
  pages     = {205--217},
  publisher = {IEEE},
  doi       = {10.1109/RTSS49844.2020.00028},
  url       = {https://par.nsf.gov/servlets/purl/10294493}
}

@inproceedings{chen2024internvl,
  title={Internvl: Scaling up vision foundation models and aligning for generic visual-linguistic tasks},
  author={Chen, Zhe and Wu, Jiannan and Wang, Wenhai and Su, Weijie and Chen, Guo and Xing, Sen and Zhong, Muyan and Zhang, Qinglong and Zhu, Xizhou and Lu, Lewei and others},
  booktitle={Proceedings of the IEEE/CVF conference on computer vision and pattern recognition},
  pages={24185--24198},
  year={2024}
}

@inproceedings{zhang2024fast,
  author    = {Lin Zhang and Luis Burbano and Xin Chen and Alvaro A. Cardenas and Steven Drager and Matthew Anderson and Fanxin Kong},
  title     = {Fast Attack Recovery for Stochastic Cyber-Physical Systems},
  booktitle = {2024 IEEE 30th Real-Time and Embedded Technology and Applications Symposium (RTAS)},
  year      = {2024},
  pages     = {280--293},
  publisher = {IEEE},
  doi       = {10.1109/RTAS61025.2024.00030}
}

@inproceedings{dash2021pidpiper,
  author    = {Pritam Dash and Guanpeng Li and Zitao Chen and Mehdi Karimibiuki and Karthik Pattabiraman},
  title     = {PID-Piper: Recovering Robotic Vehicles from Physical Attacks},
  booktitle = {2021 IEEE/IFIP International Conference on Dependable Systems and Networks (DSN)},
  year      = {2021},
  pages     = {26--38},
  publisher = {IEEE},
  url       = {https://people.ece.ubc.ca/zitaoc/files/Pid-Piper-DSN21.pdf}
}

@inproceedings{jang2023ndss,
  author    = {Joonha Jang and Mangi Cho and Jaehoon Kim and Dongkwan Kim and Yongdae Kim},
  title     = {Paralyzing Drones via {EMI} Signal Injection on Sensory Communication Channels},
  booktitle = {Network and Distributed System Security Symposium (NDSS)},
  year      = {2023},
  publisher = {Internet Society},
  doi       = {10.14722/ndss.2023.24616},
  url       = {https://cybershafarat.com/wp-content/uploads/2023/06/ParalyzingDronesviaEMISignalInjection.pdf}
}

@incollection{cardenas2025response,
  title={Response and Reconfiguration Under Attacks in CPS},
  author={Cardenas, Alvaro A},
  booktitle={Encyclopedia of Cryptography, Security and Privacy},
  pages={2100--2105},
  year={2025},
  publisher={Springer}
}

@article{barbosa2025robust,
  title={Robust and Efficient AI-Based Attack Recovery in Autonomous Drones},
  author={Barbosa, Diego Ortiz and Burbano, Luis and Yang, Siwei and Wang, Zijun and Cardenas, Alvaro A and Xie, Cihang and Cao, Yinzhi},
  journal={arXiv preprint arXiv:2505.14835},
  year={2025}
}

@inproceedings{leudo2023observer,
  title={An Observer-based Switching Algorithm for Safety under Sensor Denial-of-Service Attacks},
  author={Leudo, Santiago Jimenez and Garg, Kunal and Sanfelice, Ricardo G and Cardenas, Alvaro A},
  booktitle={2023 American Control Conference (ACC)},
  pages={2469--2474},
  year={2023},
  organization={IEEE}
}

@misc{wei2023chainofthoughtpromptingelicitsreasoning,
      title={Chain-of-Thought Prompting Elicits Reasoning in Large Language Models}, 
      author={Jason Wei and Xuezhi Wang and Dale Schuurmans and Maarten Bosma and Brian Ichter and Fei Xia and Ed Chi and Quoc Le and Denny Zhou},
      year={2023},
      eprint={2201.11903},
      archivePrefix={arXiv},
      primaryClass={cs.CL},
      url={https://arxiv.org/abs/2201.11903}, 
}

@INPROCEEDINGS{maturana3dcnnforlzd,
  author={Maturana, Daniel and Scherer, Sebastian},
  booktitle={2015 IEEE International Conference on Robotics and Automation (ICRA)}, 
  title={3D Convolutional Neural Networks for landing zone detection from LiDAR}, 
  year={2015},
  volume={},
  number={},
  pages={3471-3478},
  doi={10.1109/ICRA.2015.7139679}}

@INPROCEEDINGS{leelandingarearecognitiondl,
  author={Lee, Min-Fan Ricky and Nugroho, Asep and Le, Tuan-Tang and Bahrudin and Bastida, Saul Nieto},
  booktitle={2020 International Conference on Advanced Robotics and Intelligent Systems (ARIS)}, 
  title={Landing Area Recognition using Deep Learning for Unammaned Aerial Vehicles}, 
  year={2020},
  volume={},
  number={},
  pages={1-6},
  doi={10.1109/ARIS50834.2020.9205793}}

@article{team2025geminigeminirobotics,
  title={Gemini robotics: Bringing ai into the physical world},
  author={Gemini Robotics Team, A and Abeyruwan, Saminda and Ainslie, Joshua and Alayrac, Jean-Baptiste and Arenas, Montserrat Gonzalez and Armstrong, Travis and Balakrishna, Ashwin and Baruch, Robert and Bauza, Maria and Blokzijl, Michiel and others},
  journal={arXiv preprint arXiv:2503.20020},
  year={2025}
}

@misc{sinha2024realtimeanomalydetectionreactive,
      title={Real-Time Anomaly Detection and Reactive Planning with Large Language Models}, 
      author={Rohan Sinha and Amine Elhafsi and Christopher Agia and Matthew Foutter and Edward Schmerling and Marco Pavone},
      year={2024},
      eprint={2407.08735},
      archivePrefix={arXiv},
      primaryClass={cs.RO},
      url={https://arxiv.org/abs/2407.08735}, 
}

@INPROCEEDINGS{zhangrealtimedatapredictive,
  author={Zhang, Lin and Sridhar, Kaustubh and Liu, Mengyu and Lu, Pengyuan and Chen, Xin and Kong, Fanxin and Sokolsky, Oleg and Lee, Insup},
  booktitle={2023 IEEE 29th Real-Time and Embedded Technology and Applications Symposium (RTAS)}, 
  title={Real-Time Data-Predictive Attack-Recovery for Complex Cyber-Physical Systems}, 
  year={2023},
  volume={},
  number={},
  pages={209-222},
  doi={10.1109/RTAS58335.2023.00024}}

@article{zhangattackrecoverylqr,
author = {Zhang, Lin and Lu, Pengyuan and Kong, Fanxin and Chen, Xin and Sokolsky, Oleg and Lee, Insup},
title = {Real-time Attack-recovery for Cyber-physical Systems Using Linear-quadratic Regulator},
year = {2021},
issue_date = {October 2021},
publisher = {Association for Computing Machinery},
address = {New York, NY, USA},
volume = {20},
number = {5s},
issn = {1539-9087},
url = {https://doi.org/10.1145/3477010},
doi = {10.1145/3477010},
journal = {ACM Trans. Embed. Comput. Syst.},
month = sep,
articleno = {79},
numpages = {24}
}

@article{bhatt2024know,
  title={Know Where You're Uncertain When Planning with Multimodal Foundation Models: A Formal Framework},
  author={Bhatt, Neel P and Yang, Yunhao and Siva, Rohan and Milan, Daniel and Topcu, Ufuk and Wang, Zhangyang},
  journal={arXiv preprint arXiv:2411.01639},
  year={2024}
}

@misc{yang2024depthv2,
      title={Depth Anything V2}, 
      author={Lihe Yang and Bingyi Kang and Zilong Huang and Zhen Zhao and Xiaogang Xu and Jiashi Feng and Hengshuang Zhao},
      year={2024},
      eprint={2406.09414},
      archivePrefix={arXiv},
      primaryClass={cs.CV},
      url={https://arxiv.org/abs/2406.09414}, 
}

@misc{amatriain2024promptdesignengineeringintroduction,
      title={Prompt Design and Engineering: Introduction and Advanced Methods}, 
      author={Xavier Amatriain},
      year={2024},
      eprint={2401.14423},
      archivePrefix={arXiv},
      primaryClass={cs.SE},
      url={https://arxiv.org/abs/2401.14423}, 
}

@misc{sahoo2025systematicsurveypromptengineering,
      title={A Systematic Survey of Prompt Engineering in Large Language Models: Techniques and Applications}, 
      author={Pranab Sahoo and Ayush Kumar Singh and Sriparna Saha and Vinija Jain and Samrat Mondal and Aman Chadha},
      year={2025},
      eprint={2402.07927},
      archivePrefix={arXiv},
      primaryClass={cs.AI},
      url={https://arxiv.org/abs/2402.07927}, 
}

@misc{eigen2014depthmappredictionsingle,
      title={Depth Map Prediction from a Single Image using a Multi-Scale Deep Network}, 
      author={David Eigen and Christian Puhrsch and Rob Fergus},
      year={2014},
      eprint={1406.2283},
      archivePrefix={arXiv},
      primaryClass={cs.CV},
      url={https://arxiv.org/abs/1406.2283}, 
}

@Article{yangmonocularvisionslam,
AUTHOR = {Yang, Tao and Li, Peiqi and Zhang, Huiming and Li, Jing and Li, Zhi},
TITLE = {Monocular Vision SLAM-Based UAV Autonomous Landing in Emergencies and Unknown Environments},
JOURNAL = {Electronics},
VOLUME = {7},
YEAR = {2018},
NUMBER = {5},
ARTICLE-NUMBER = {73},
URL = {https://www.mdpi.com/2079-9292/7/5/73},
ISSN = {2079-9292},
DOI = {10.3390/electronics7050073}
}

@article{shahalamslzsurvey,
title = {A survey of safe landing zone detection techniques for autonomous unmanned aerial vehicles (UAVs)},
journal = {Expert Systems with Applications},
volume = {179},
pages = {115091},
year = {2021},
issn = {0957-4174},
doi = {https://doi.org/10.1016/j.eswa.2021.115091},
url = {https://www.sciencedirect.com/science/article/pii/S0957417421005327},
author = {Md {Shah Alam} and Jared Oluoch}
}

@inproceedings{garg2022control,
  title={Control barrier function-based attack-recovery with provable guarantees},
  author={Garg, Kunal and Sanfelice, Ricardo G and Cardenas, Alvaro A},
  booktitle={2022 IEEE 61st Conference on Decision and Control (CDC)},
  pages={4808--4813},
  year={2022},
  organization={IEEE}
}

@inproceedings{mbv2025cvpr,
 author = {Byrd Victorica, Maruicio and D\'{a}n, Gy\"{o}rgy and Sandberg, Henrik},
 title = {Saliuitl: Ensemble Salience Guided Recovery of Adversarial Patches against {CNNs}},
 booktitle = {IEEE/CVF CVPR},
 year = {2025},
}

@article{halton1960efficiency,
  title={On the efficiency of certain quasi-random sequences of points in evaluating multi-dimensional integrals},
  author={Halton, John H},
  journal={Numerische Mathematik},
  volume={2},
  number={1},
  pages={84--90},
  year={1960},
  publisher={Springer}
}

@article{dreossi2019compositional,
  title={Compositional falsification of cyber-physical systems with machine learning components},
  author={Dreossi, Tommaso and Donz{\'e}, Alexandre and Seshia, Sanjit A},
  journal={Journal of Automated Reasoning},
  volume={63},
  number={4},
  pages={1031--1053},
  year={2019},
  publisher={Springer}
}

@inproceedings{dash2024specguard,
author = {Dash, Pritam and Chan, Ethan and Pattabiraman, Karthik},
title = {SpecGuard: Specification Aware Recovery for Robotic Autonomous Vehicles from Physical Attacks},
year = {2024},
isbn = {9798400706363},
publisher = {Association for Computing Machinery},
address = {New York, NY, USA},
url = {https://doi.org/10.1145/3658644.3690210},
doi = {10.1145/3658644.3690210},
booktitle = {Proceedings of the 2024 on ACM SIGSAC Conference on Computer and Communications Security},
pages = {1849–1863},
numpages = {15},
location = {Salt Lake City, UT, USA},
series = {CCS '24}
}

@INPROCEEDINGS{roque2022corridormpc,
  author={Roque, P. and Cortez, W. Shaw and Lindemann, L. and Dimarogonas, D. V.},
  booktitle={2022 American Control Conference (ACC)}, 
  title={Corridor MPC: Towards Optimal and Safe Trajectory Tracking}, 
  year={2022},
  volume={},
  number={},
  pages={2025-2032},
  doi={10.23919/ACC53348.2022.9867764}}

@InProceedings{prakash2021multimodaltransformer,
    author    = {Prakash, Aditya and Chitta, Kashyap and Geiger, Andreas},
    title     = {Multi-Modal Fusion Transformer for End-to-End Autonomous Driving},
    booktitle = {Proceedings of the IEEE/CVF Conference on Computer Vision and Pattern Recognition (CVPR)},
    month     = {June},
    year      = {2021},
    pages     = {7077-7087}
}

@ARTICLE{xiao2022endtoenddriving,
  author={Xiao, Yi and Codevilla, Felipe and Gurram, Akhil and Urfalioglu, Onay and López, Antonio M.},
  journal={IEEE Transactions on Intelligent Transportation Systems}, 
  title={Multimodal End-to-End Autonomous Driving}, 
  year={2022},
  volume={23},
  number={1},
  pages={537-547},
  doi={10.1109/TITS.2020.3013234}}

@article{Jolliffe2016PrincipalCA,
  title={Principal component analysis: a review and recent developments},
  author={Ian T. Jolliffe and Jorge Cadima},
  journal={Philosophical Transactions of the Royal Society A: Mathematical, Physical and Engineering Sciences},
  year={2016},
  volume={374},
  url={https://api.semanticscholar.org/CorpusID:20101754}
}

@article{10.1007/s11263-009-0275-4,
author = {Everingham, Mark and Gool, Luc and Williams, Christopher K. and Winn, John and Zisserman, Andrew},
title = {The Pascal Visual Object Classes (VOC) Challenge},
year = {2010},
issue_date = {June      2010},
publisher = {Kluwer Academic Publishers},
address = {USA},
volume = {88},
number = {2},
issn = {0920-5691},
url = {https://doi.org/10.1007/s11263-009-0275-4},
doi = {10.1007/s11263-009-0275-4},
journal = {Int. J. Comput. Vision},
month = jun,
pages = {303–338},
numpages = {36}
}

@misc{delatorrevanegas2025visionbasedriskawareemergency,
      title={Vision-Based Risk Aware Emergency Landing for UAVs in Complex Urban Environments}, 
      author={Julio de la Torre-Vanegas and Miguel Soriano-Garcia and Israel Becerra and Diego Mercado-Ravell},
      year={2025},
      eprint={2505.20423},
      archivePrefix={arXiv},
      primaryClass={cs.RO},
      url={https://arxiv.org/abs/2505.20423}, 
}

@InProceedings{safeuavmarcualina,
author="Marcu, Alina
and Costea, Drago{\c{s}}
and Lic{\u{a}}re{\c{T}}, Vlad
and P{\^i}rvu, Mihai
and Slu{\c{s}}anschi, Emil
and Leordeanu, Marius",
editor="Leal-Taix{\'e}, Laura
and Roth, Stefan",
title="SafeUAV: Learning to Estimate Depth and Safe Landing Areas for UAVs from Synthetic Data",
booktitle="Computer Vision -- ECCV 2018 Workshops",
year="2019",
publisher="Springer International Publishing",
address="Cham",
pages="43--58",
isbn="978-3-030-11012-3"
}

@article{SAM,
  title={Segment Anything},
  author={Kirillov, Alexander and Mintun, Eric and Ravi, Nikhila and Mao, Hanzi and Rolland, Chloe and Gustafson, Laura and Xiao, Tete and Whitehead, Spencer and Berg, Alexander C. and Lo, Wan-Yen and Doll{\'a}r, Piotr and Girshick, Ross},
  journal={arXiv:2304.02643},
  year={2023}
}

@misc{OpenLander,
  author = {Sturges, Stephan},
  title = {OpenLander: free-to-use landing-zone detection for {UAVs}},
  year = {2022},
  publisher = {GitHub},
  journal = {GitHub repository},
  howpublished = {\url{https://github.com/stephansturges/OpenLander}}
}


\end{document}